\documentclass[10pt,twocolumn,letterpaper]{article}

\usepackage{cvpr}      %
\usepackage{graphicx}
\usepackage{amsmath}
\usepackage{amssymb}
\usepackage{booktabs}
\usepackage{color}
\usepackage{makecell, tabularx}
\usepackage{rotating}
\usepackage[export]{adjustbox}
\usepackage[pagebackref,breaklinks,colorlinks]{hyperref}

\def\etal{\emph{et al}\onedot}

\usepackage[capitalize]{cleveref}
\crefname{section}{Sec.}{Secs.}
\Crefname{section}{Section}{Sections}
\Crefname{table}{Table}{Tables}
\crefname{table}{Tab.}{Tabs.}

\begin{document}

\title{RelightableHands: Efficient Neural Relighting of Articulated Hand Models}

\author{
Shun Iwase$^{1,2*}$ \qquad 
Shunsuke Saito$^2$ \qquad
Tomas Simon$^2$ \qquad \\
Stephen Lombardi$^2$ \qquad 
Timur Bagautdinov$^2$ \qquad
Rohan Joshi$^2$ \qquad \\
Fabian Prada$^2$ \qquad
Takaaki Shiratori$^2$ \qquad 
Yaser Sheikh$^2$ \qquad
Jason Saragih$^2$ \qquad
\\$^1$Carnegie Mellon University \qquad $^2$Reality Labs Research \\
}
\twocolumn[{%
\renewcommand\twocolumn[1][]{#1}%
\maketitle
\begin{center}
    \centering
    \begin{minipage}{0.16\textwidth}
       \centering
       \includegraphics[trim={2cm 20cm 18cm 15cm},clip,width=1.0\textwidth]{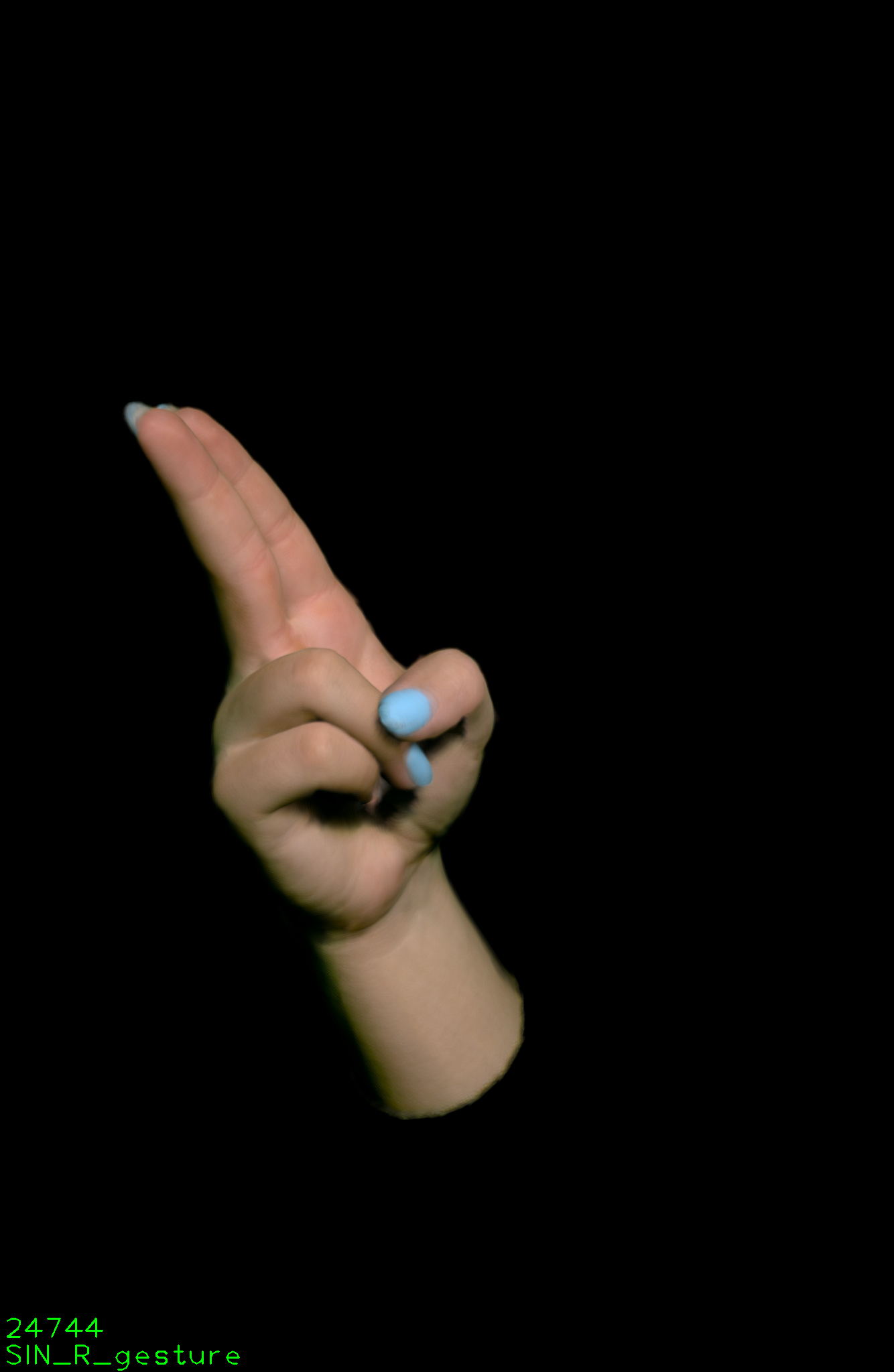}
       \label{fig:my_label}
        \vspace{-0.8cm}
        \captionof*{figure}{\scriptsize{Directional Light}}
    \end{minipage}
    \begin{minipage}{0.16\textwidth}
       \centering
       \includegraphics[trim={0cm 18cm 21.5cm 19cm},clip,width=1.0\textwidth]{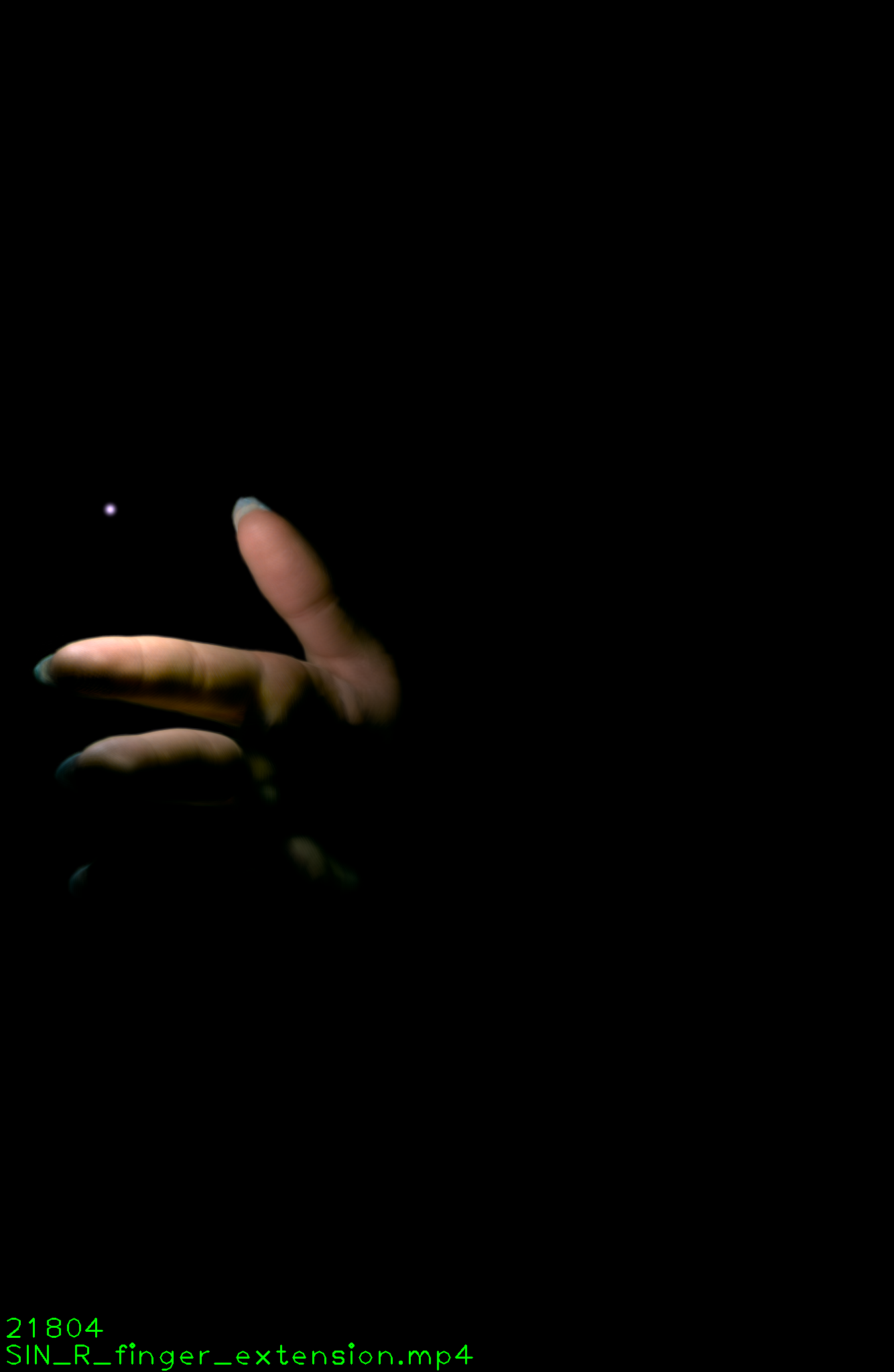}
        \label{fig:my_label}
        \vspace{-0.8cm}
        \captionof*{figure}{\scriptsize{Near-field Light}}
    \end{minipage}
    \begin{minipage}{0.16\textwidth}
       \centering
       \includegraphics[trim={9.7cm 24cm 14cm 16cm},clip,width=1.0\textwidth]{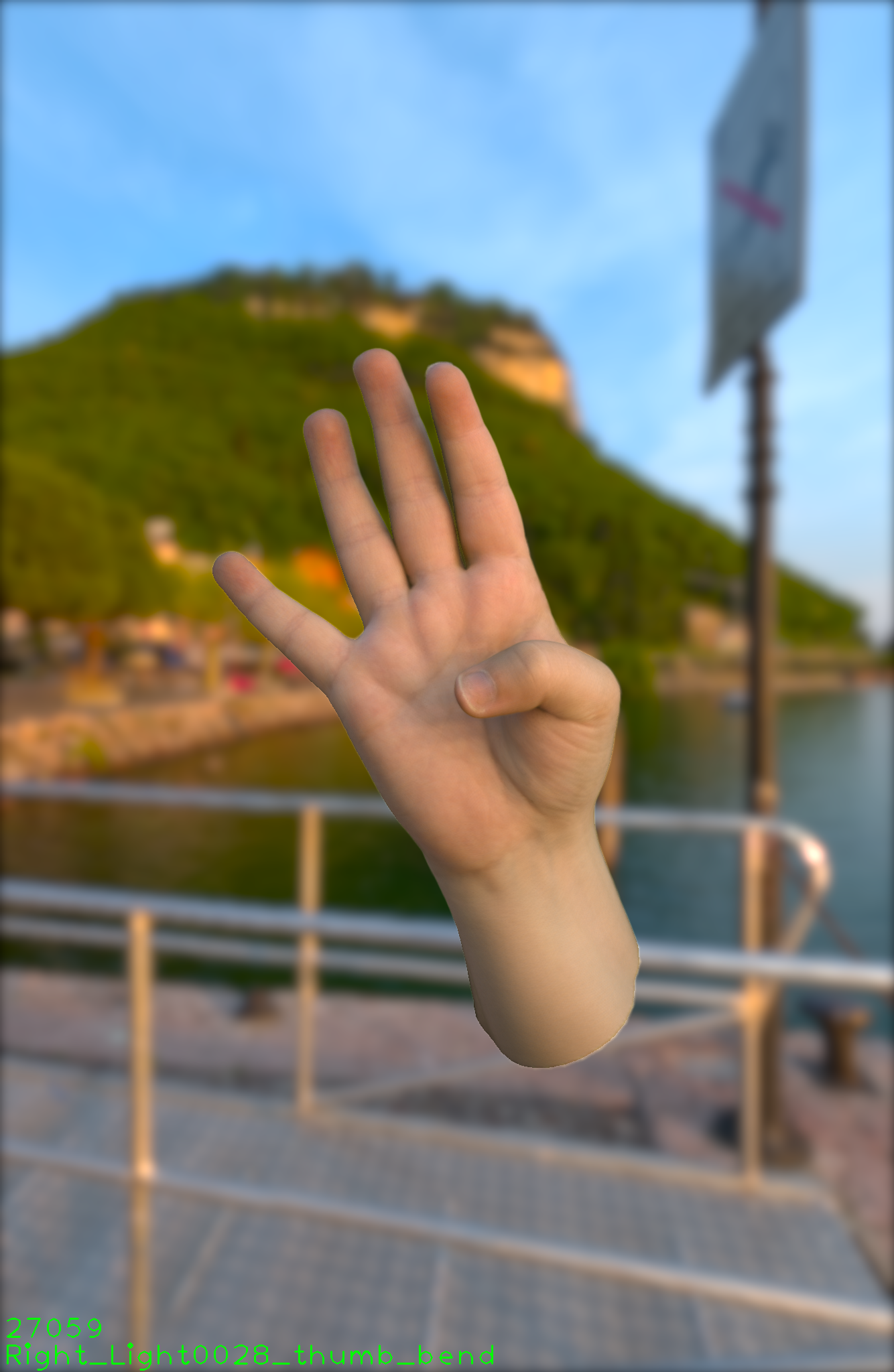}
        \vspace{-0.8cm}
        \captionof*{figure}{\scriptsize{Environment Map}}
       \label{fig:my_label}
    \end{minipage}
    \begin{minipage}{0.32\textwidth}
       \centering
       \includegraphics[width=0.991\textwidth]{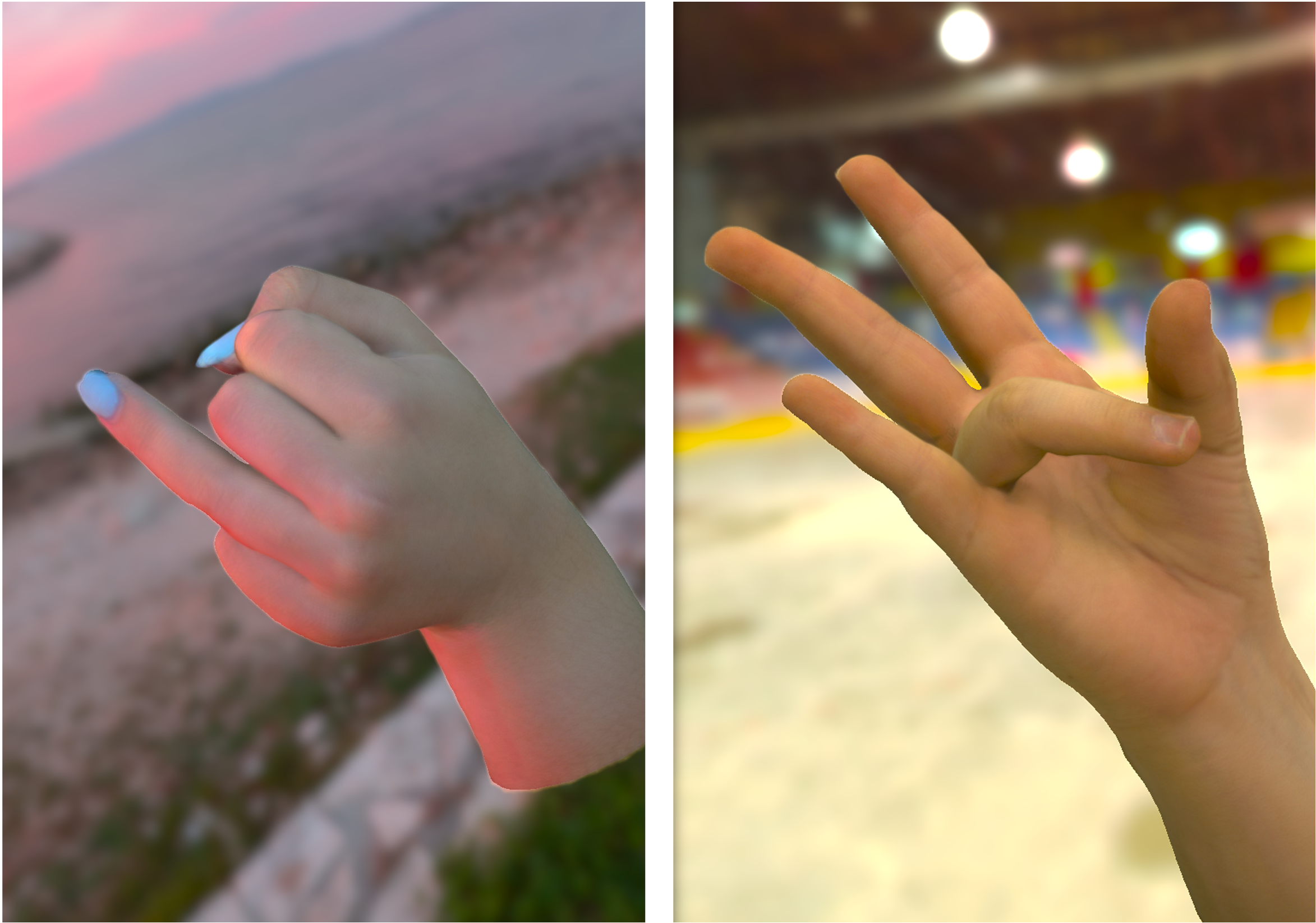}
       \label{fig:my_label}
    \vspace{-0.8cm}
    \captionof*{figure}{\scriptsize{Student Model Rendering (21ms)}}
    \end{minipage}
    \begin{minipage}{0.17\textwidth}
       \centering
       \includegraphics[trim={4cm 4cm 13cm 29cm},clip,width=1.0\textwidth]{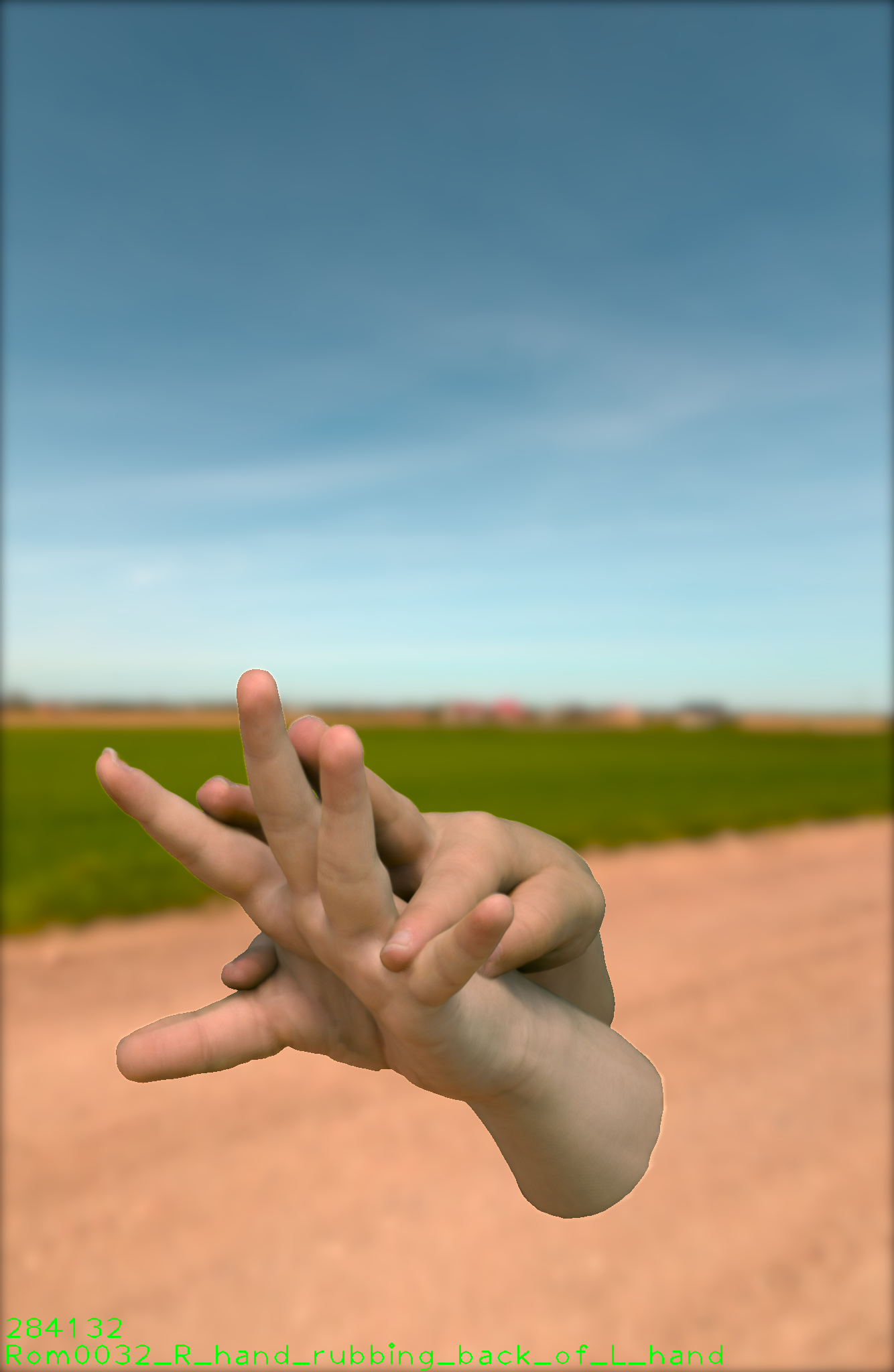}
       \label{fig:my_label}
    \vspace{-0.8cm}
    \captionof*{figure}{\scriptsize{Two Hand Rendering (32ms)}}
    \end{minipage}
    \vspace{-0.2cm}
    \captionof{figure}{
\textbf{Neural Relighting of Animatable Hands.} Our model-based neural rendering approach enables high-fidelity rendering of hands with various poses, views, and illuminations. Our student model is highly efficient enough to render in real-time.
}
    \label{fig:teaser}
\end{center}%
}]

\let\thefootnote\relax\footnotetext{$^*$This work was done during an internship at Meta}

\begin{abstract}

We present the first neural relighting approach for rendering high-fidelity personalized hands that can be animated in real-time under novel illumination. 
Our approach adopts a teacher-student framework, where the teacher learns appearance under a single point light from images captured in a light-stage, allowing us 
to synthesize hands in arbitrary illuminations but with heavy compute.
Using images rendered by the teacher model as training data, an efficient student model directly predicts appearance under natural illuminations in real-time.
To achieve generalization, we condition the student model with physics-inspired illumination features such as visibility, diffuse shading, and specular reflections computed on a coarse proxy geometry, maintaining a small computational overhead.
Our key insight is that these features have strong correlation with subsequent global light transport effects, which proves sufficient as conditioning data for the neural relighting network. 
Moreover, in contrast to bottleneck illumination conditioning, these features are spatially aligned based on underlying geometry, leading to better generalization to unseen illuminations and poses. 
In our experiments, we demonstrate the efficacy of our illumination feature representations, outperforming baseline approaches.
We also show that our approach can photorealistically relight two interacting hands at real-time speeds.
\href{https://sh8.io/#/relightable_hands}{https://sh8.io/\#/relightable\_hands}
\end{abstract}

\section{Introduction}
\label{sec:intro}

Neural rendering approaches have 
significantly advanced
photorealistic face rendering~\cite{sun2019single,Pandey:2021,yeh2022learning}
in recent years.
These methods use deep neural networks to model the light transport on human skin~\cite{weyrich2006analysis,Ma07,Ghosh11,Fyffe14}, 
directly reproducing physical effects such as subsurface scattering by 
reconstructing real
images.
However, despite the success of neural relighting, extending this approach to animatable hand models poses a unique 
challenge: generalization across articulations.

Unlike faces, hands have many joints, and the 
state
of a single joint affects all child joints.
This leads to extremely diverse shape variations even within a single subject.
Changes in pose drastically affect the appearance of hands, creating wrinkles, casting shadows, and inter-reflecting across topologically distant regions.
Rendering these effects is challenging because sufficiently accurate geometry and material properties required for photorealism are difficult to obtain, and even then, path tracing to sufficient accuracy is computationally expensive.
The use of simplified geometric and appearance models (such as linear blend skinning and reduced material models) allow faster computation but come at a noticeable 
degradation
in rendering fidelity.
So far, 
photorealistic rendering of animatable hands with global illumination effects in real-time remains an open problem.

In this work, we aim to enable photorealistic rendering of a personalized hand model that can be animated with novel poses, in novel lighting environments, and supports rendering two-hand interactions.
To this end, we present the first neural relighting framework of a parameteric 3D hand model for real-time rendering.
Specifically, we build a relightable hand model to reproduce light-stage captures of dynamic hand motions. 

Inspired by~\cite{bi2021deep}, we capture performances under spatiotemporal-multiplexed illumination patterns, where 
fully-on illumination is 
interleaved to 
enable tracking of the 
current state of hand geometry and poses. We use a two-stage teacher-student approach to learn a model that generalizes to natural illuminations outside of the capture system. 
We first train a teacher model that infers radiance given a point-light position, a viewing direction, and light visibility.
As this model directly learns the mapping between an input light position and output radiance, it can accurately model complex reflectance and scattering on the hand without the need for path tracing.
To render hands in arbitrary illuminations, we treat natural illuminations as a combination of distant point-light sources %
by using the linearity of light transport~\cite{debevec2000acquiring}. 
We then take renderings from the teacher model as pseudo ground-truth to train an efficient student model that is conditioned on the target environment maps.

However, we found that the student model architecture used in~\cite{bi2021deep} for faces leads to severe overfitting when applied to relightable hands. This is caused by the architecture 
design of holistically conditioning a bottleneck representation with the target lighting environment.
This representation makes it difficult to reproduce geometric interactions between lights and hand pose, such as those required to cast shadows from the fingers onto the palm across all possible finger configurations.

Therefore, motivated by recent neural portrait relighting works~\cite{wang2020single,Pandey:2021}, we instead propose to compute spatially aligned lighting information using physics-inspired illumination features, including visibility, diffuse shading, and specular reflections.
Because these features are based on geometry and approximate the first bounce of light transport, they show strong correlation with the full appearance and provide sufficient conditioning information to infer accurate radiance under natural illuminations. 
In particular, visibility plays a key role in disentangling lights and pose, reducing the learning of spurious correlations that can be present in limited training data.
However, computing visibility at full geometric resolution for every single light is too computationally expensive for real-time rendering.
To address this, we propose using a coarse proxy mesh that shares the same UV parameterization as our hand model for computing the lighting features.
We compute the features at vertices of the coarse geometry, and use barycentric interpolation to create texel-aligned lighting features.
Our fully convolutional architecture learns to compensate for the approximate nature of the input features and infers both local and global 
light transport effects. This way, our model can render appearance under natural illuminations at real-time framerates as shown in Figure~\ref{fig:teaser}. 

Our study shows that both integrating visibility information and spatially aligned illumination features are important for generalization to novel illuminations and poses. We also demonstrate that our approach supports rendering of two hands in real-time, with realistic shadows cast across hands.

Our contributions can be summarized as follows:
\begin{itemize}
    \item The first method to learn a relightable personalized hand model from multi-view light-stage data that supports high-fidelity relighting under novel lighting environments.
    \item An illumination representation for parametric model relighting that is spatially aligned, leading to significant improvements in generalization and accuracy of shadows under articulation.
    \item An efficient algorithm to compute spatially-aligned lighting features with visibility and shading information incorporated using a coarse proxy mesh, enabling real-time synthesis.
\end{itemize}

\section{Related Work}

In the following, we review image-space and model-based relighting approaches as well as hand modeling techniques.

\smallskip \noindent {\bf Hand Modeling}
Modeling human hands has been extensively studied in both computer vision and graphics. 
Early work primarily focuses on tracking geometry and modeling articulation. Various hand shape representations have been proposed including simple shape primitives~\cite{rehg1994visual,oikonomidis2011efficient,tagliasacchi2015robust}, sum of 3D Gaussians~\cite{sridhar2013interactive,sridhar2015fast}, sphere mesh~\cite{tkach2016sphere}, and triangle meshes~\cite{de2011model,ballan2012motion,tzionas2016capturing,MANO:SIGGRAPHASIA:2017}. 
MANO~\cite{MANO:SIGGRAPHASIA:2017} presents a parametric mesh model that learns identity variations as well as pose dependent deformations. 
Facilitated by such parametric models and accurate joint detection methods~\cite{simon2017hand}, estimating 3D hand poses is now possible from RGB-D inputs~\cite{mueller2017real,moon2018v2v} or images~\cite{mueller2018ganerated,moon2020i2l,zhou2020monocular}. 
They are also extended to two hands~\cite{mueller2019real,li2022interacting} and object interactions~\cite{hasson2019learning}. 
While these approaches show impressive robustness, geometric fidelity remains limited.
To further improve fidelity of geometry modeling, anatomical priors from medical images~\cite{wang2019hand,li2022nimble,Zheng:2022:SOH} and physics-based volumetric prior~\cite{smith2020constraining} allows modeling more accurate surface deformation, especially around articulation and contacts.
Self-supervised learning enables the learning of more personalized articulated models in an end-to-end manner with a mesh representation~\cite{moon2020deephandmesh} and neural fields~\cite{deng2019neural,saito2021scanimate,karunratanakul2021halo,alldieck2021imghum}.

The appearance of hands is also essential for realistic animation.
HTML~\cite{HTML_eccv2020} builds a database of hand textures to create a parameteric texture space that can be fit to novel hands.
Neural rendering approaches based on volumetric representation have been extended to articulation modeling, compensating for inaccurate geometry by using view-dependent appearance~\cite{peng2021neural,2021narf}.
In particular, LISA~\cite{corona2022lisa} demonstrates the modeling of animatable hands from multi-view images.
However, these approaches pre-integrate illuminations into the appearance model, and relighting is not supported.
NIMBLE~\cite{li2022nimble} captures diffuse, normal, and specular maps from a light-stage and build a PCA appearance space.
While reflactance maps allow relighting, physically-based rendering requires
expensive raytracing and is sensitive to geometry quality, while linear appearance models have limited capacity for compensating geometry errors.

In contrast, our work proposes an end-to-end model for geometry and relightable appearance by leveraging neural rendering techniques. By directly reproducing complex light transport effects using neural rendering, our method can achieve significantly more efficient photorealistic relighting.

\smallskip \noindent {\bf Image-space Human Relighting}
Image-space relighting has been pioneered by Devebec~\etal~\cite{debevec2000acquiring}, where faces under novel illuminations are generated by making use of the linearity of light transport from a one-light-at-a-time (OLAT) capture.
A follow-up work by Wenger \etal~\cite{wenger2005performance} enables dynamic relighting by warping adjacent frames with time-multiplexed illumination patterns. 
The learning-based approach of Xu \etal~\cite{xu2018deep} proposes to interpolate light positions from sparse observations, and a similar approach is extended to light-stage captures by upsampling light directions~\cite{sun2020light}.
Meka \etal~\cite{meka2019deep} also infer OLAT images from a pair of spherical gradient illuminations, enabling dynamic captures.
In contrast to these approaches based on single point lights, Sun \etal~\cite{sun2019single} directly regress faces under natural illuminations using deep neural networks. 
A parallel line of work aims to decompose images into geometry and reflectances, enabling physically based relighting~\cite{face-relighting-with-geometrically-consistent-shadows,DBLP:conf/cvpr/NestmeyerLML20,towards-high-fidelity-face-relighting-with-realistic-shadows,sfsnetSengupta18,kanamori2018relight,Lagunas:2021}.
Recent works leverage the best of learning-based relighting and material decomposition by feeding physics-inspired relighting results from the estimation into another network to produce a final relighting image~\cite{wang2020single,Pandey:2021,Ji_undated-vp,yeh2022learning}. 
Despite plausible relighting results, image-space approaches typically suffer from temporal and view inconsistency artifacts during animation or novel-view rendering, as they lack a 3D 
parameterization of the scene.

\smallskip \noindent {\bf Model-based Human Relighting}
In contrast to image-space neural relighting approaches, we can leverage a 3D template-model for animation and novel-view rendering. 
Yamaguchi \etal~\cite{yamaguchi2018high} infer skin reflectance from a single image in a shared UV space, allowing them to relight faces from different views. 
Zhang \etal~\cite{zhang2021nlt} also leverage a shared UV space to relight novel-views of human performance captures with global light transport. 
Unfortunately, this approach only supports a playback of existing performances and cannot create new animations. 
To enable animatable relighting for facial performance, Bi~\etal~\cite{bi2021deep} presents DRAM, a deep relightable appearance model that is conditioned on viewing direction and expression latent codes.
While their approach enables efficient relighting for real-time animation using a teacher-student framework, we observe that the bottleneck lighting encoding without visibility information in their student model leads to severe overfitting when applied to hand relighting.
EyeNeRF~\cite{li2022eyenerf} enables the joint learning of geometry and relightable appearance of a moving eyeball model.
Compared to eyes, hands exhibit significantly more diverse pose variations, making explicit visibility incorporation essential.
Relighting4D~\cite{chen2022relighting} learns relightable materials of an articulated human under a single unknown illumination, but the fidelity of relighting is limited bu the expressiveness of their parametric BRDF model.
In contrast to these methods, our approach enables relighting of articulate hand models that can be animated with 
a wide range of
poses. 
In addition, the proposed lighting encoding makes our relightable model generalizable to novel poses and illuminations while retaining real-time performance.

\section{Preliminaries}
\label{sec:pre}

\paragraph{Data Acquisition.}

We use a multiview calibrated capture system consisting of $106$ cameras and $460$ white LED lights to capture both fully-lit and partially-lit images of hands in motion, using a setup similar to~\cite{bi2021deep}. Images are captured at $4096\times2668$ resolution at $90$ frames per second. We represent the state of a hand using pose parameters, and estimate them for all the frames in the following way:
for fully-lit frames, we perform skeletal hand tracking using a personalized Linear Blend Skinning (LBS) model. 
Specifically, we first obtain 3D reconstruction meshes using~\cite{guo2019relightables} and detect 3D hand keypoints using~\cite{li2019rethinking} with a ResNet backbone and RANSAC-based triangulation.
An LBS model is personalized using reconstructions and keypoints on a collection of key frames, and is used for skeletal tracking~\cite{gall2009motion} to estimate pose parameters for fully-lit frames.
For partially-lit frames, 
we perform spherical linear interpolation of the pose parameters from adjacent fully-lit frames.
Our dataset contains independently captured sequences of right and left hands. We collected $92,313$ and $88,413$ frames for Subject 1's hands and 
$22,754$ and $22,354$ frames for Subject 2 from $106$ cameras. $80$\% of the segments are used for training and the rest for testing.

\begin{figure*}[t!]
    \centering
    \includegraphics[width=0.9\textwidth]{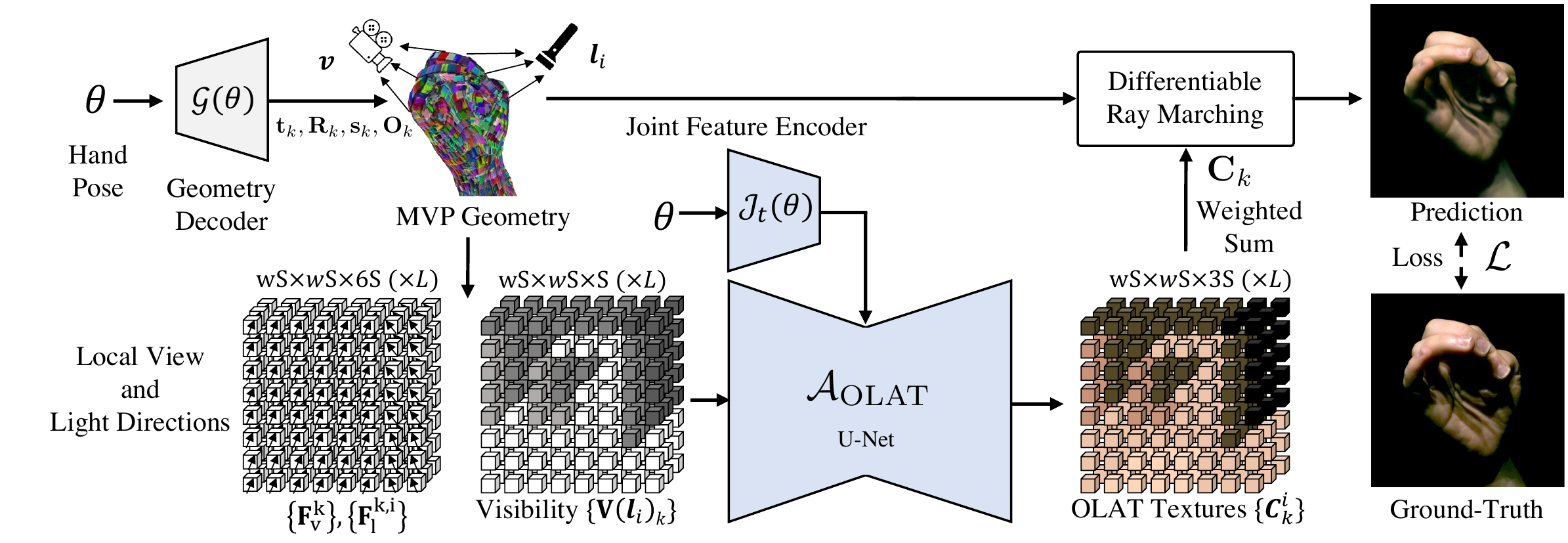}
    \caption{\textbf{Overview of the teacher model.} Our texture decoder U-Net takes as input view and light directions in local primitive coordinates, and visibility at each voxel in the primitives. Input pose parameters are encoded into joint features, and fed into the bottleneck layer of U-Net. The output OLAT textures are aggregated by the weighted sum using the intensity of each light. The network weights are trained in an end-to-end manner via inverse rendering losses.}
    \label{fig:overview_teacher}
    \vspace{-0.5cm}
\end{figure*}

\paragraph{Articulated Geometry Modeling.}
\label{par:geometry_rep}
We adopt the articulated mixture of volumetric primitives (MVP)~\cite{remelli2022dva}, which extends the original work of MVP~\cite{lombardi2021mvp} to articulated objects. As demonstrated in~\cite{remelli2022dva}, the articulated MVP improves fidelity over mesh-based representations due to its volumetric nature while being computationally efficient for real-time rendering. Additionally, it only requires a coarse mesh from an LBS model as guidance, in contrast to prior mesh-based works~\cite{lombardi2018deep,bagautdinov2021driving}, which rely on precise surface tracking.

Given a coarse articulated mesh $\mathcal{M}=\{ \mathcal{V},\mathcal{T}, \mathcal{F}, \theta \}$ with vertices $\mathcal{V}$, texture coordinates $\mathcal{T}$, faces $\mathcal{F}$, and hand pose parameters $\theta$ representing joint angles, we decode a set of volumetric primitives. 
Specifically, our pose-dependent hand geometry is modeled by $N$ primitives, where the $k$-th primitive is defined by $\mathcal{P}_k =\left\{\mathbf{t}_k, \mathbf{R}_k, \mathbf{s}_k, \mathbf{C}_k, \mathbf{O}_k\right\}$, comprising the primitive center location $\mathbf{t}_k \in \mathbb{R}^{3}$, rotation $\mathbf{R}_k \in SO(3)$, per-axis scale $s_k \in \mathbb{R}^{3}$, and voxels that contain color $\mathbf{C}_k \in \mathbb{R}^{3 \times S \times S \times S}$ and opacity $\mathbf{O}_k \in \mathbb{R}^{S \times S \times S}$ for each primitive, where $S$ denotes the resolution of voxels on each axis. 
To explicitly model articulations, primitives are loosely attached to the articulated mesh $\mathcal{M}$ produced by LBS. 
Given pose $\theta$, the geometry decoder $\mathcal{G}(\theta)$ predicts residual rotations, translations, and scale together with the opacity of primitives $\{\mathbf{O}_k\}_{k=1}^N$. 
The texture decoder $\mathcal{C}(\theta)$ predicts the color of primitives $\{\mathbf{C}_k\}_{k=1}^N$. 
Both color and opacity decoders employ a sequence of 2D transpose convolutions, 
and the channel dimension in the last layer additionally stacks the depth-axis of each primitive's voxels. 
The decoded primitives $\{\mathcal{P}_k\}_{k=1}^N$ are rendered using differentiable ray marching~\cite{lombardi2021mvp};
we refer to~\cite{remelli2022dva} for details.

In this work, we first train the articulated MVP~\cite{remelli2022dva} from fully-lit images without relighting to obtain a personalized geometry decoder $\mathcal{G}(\theta)$. 
After training, we discard the non-relightable texture decoder $\mathcal{C}(\theta)$ and learn a relightable appearance decoder. 

\section{Method}

Our goal is to build a relightable appearance model for hands that can be rendered under natural illuminations in real-time from a light-stage capture based on point lights. To this end, we use a similar teacher-student framework as proposed in~\cite{bi2021deep}, but extend it to articulated MVPs. The teacher model learns OLAT relightable textures using the partially-lit frames. Because the teacher model computes illumination for single point light sources, it generalizes to arbitrary illuminations due to the linearity of light transport~\cite{debevec2000acquiring}. However, multiple OLAT textures need to be generated to obtain a rendering under natural illumination. This leads to significant computational overheads for rendering (${\sim}30$s per frame). Thus, we use the teacher OLAT model to synthesize images under natural illuminations, and train an efficient student model that can be conditioned by an environment map to match with the pseudo ground-truth generated by the teacher model.

\subsection{Teacher Model}
\label{sec:teacher}

\begin{figure*}[t!]
    \centering
    \includegraphics[width=0.9\textwidth]{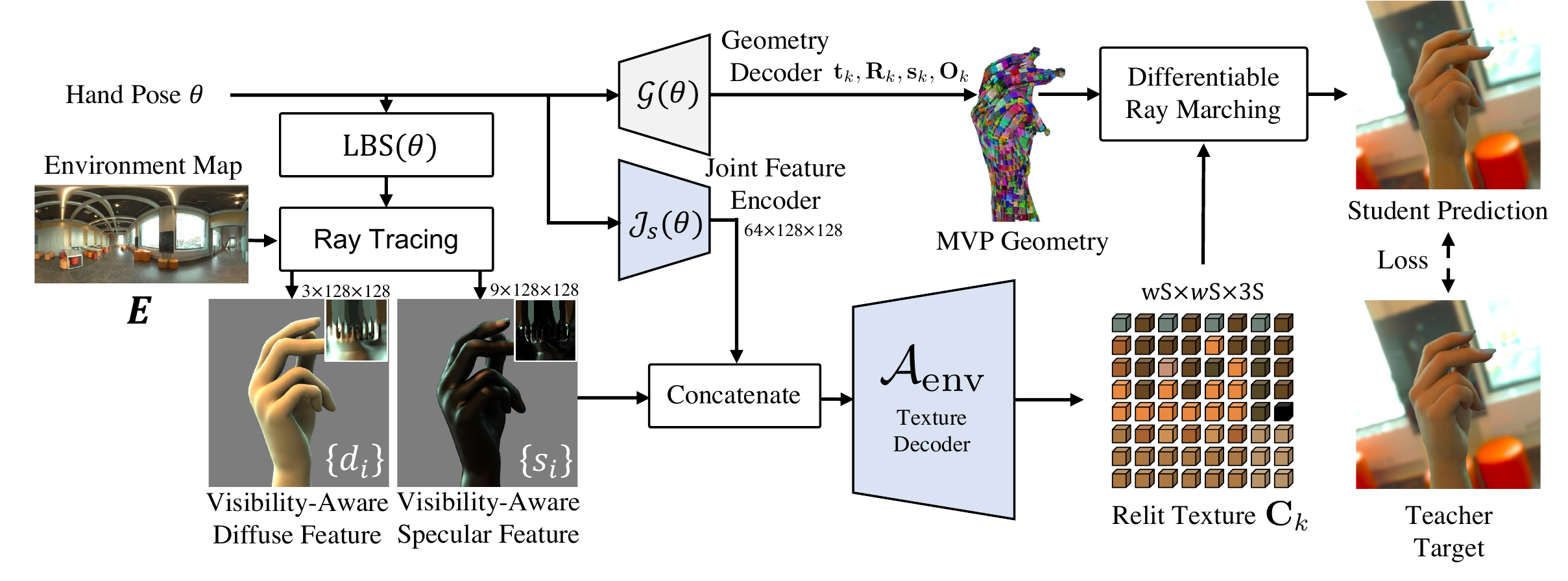}
    \vspace{-0.5cm}
    \caption{\textbf{Overview of the student model.} Given a hand pose and a target envmap, the visibility-aware diffuse and specular features are computed on the coarse LBS mesh. These features are then projected onto UV map and fed into the texture decoder together with the joint encoding. Finally, the predicted texture is rendered to image space via differentiable ray marching for supervision. }
    \label{fig:overview_student}
    \vspace{-0.5cm}
\end{figure*}

Our teacher model $\mathcal{A}_{\text{OLAT}}$ predicts the appearance of the hand model under OLAT as follows:
\begin{equation}
    \{\mathbf{C}^i_k\}_{k} = \mathcal{A}_{\text{OLAT}}(\theta, \mathbf{v}, \mathbf{l}_i, \{\mathbf{V}(\mathbf{l}_i)_k\}_{k}),
\end{equation}
where $\mathbf{v}$ is the viewer's position, $\mathbf{l}_i$ is the position of $i$-th point light, and $\mathbf{V}(\mathbf{l}_i)_k {\in} \mathbb{R}^{S \times S \times S}$ are the visibility maps from the primitive to light $\mathbf{l}_i$ computed using Deep Shadow Maps~\cite{lokovic2000deep}\footnote{Raymarching from each light to primitives and accumulating opacity.}. 
Instead of OLAT, our partially lit frames use $L{=}5$ grouped lights to increase brightness and reduce motion blur. By leveraging the linearity of light transport, the final color $\{\mathbf{C}^i_k\}$ under these lights is computed as the weighted sum of primitive colors for each light:
\begin{equation}
    \vspace{-0.2cm}
    \mathbf{C}_k = \sum_{i=1}^L{b_i \mathbf{C}^i_k},
\end{equation}
where $b_i$ is the intensity of each light and $L$ is the total number of lights.

Compared to a mesh-based OLAT teacher model for face relighting~\cite{bi2021deep}, we made several important modifications in the architecture design to support hand relighting with a hybrid mesh-volumetric representation. 
In~\cite{bi2021deep}, view-dependent intrinsic feature maps are generated at $512\times512$ resolution, and per-texel features are further transformed by an MLP together with the incoming light direction to infer radiance. 
However, an MVP-based decoder requires significantly larger channel dimensions to represent the additional volumetric depth axis. 
Hence, decoding radiance at every single voxel using an MLP is not computationally tractable. 
To address this, we adopt a U-Net architecture that takes as input reshaped visibility maps and spatially aligned light and view directions. Namely, for each primitive~$k$ and light~$i$, light directions are encoded as $\mathbf{F}^{k,i}_l \in \mathbb{R}^{3 \times S \times S \times S}$ and viewing directions as $\mathbf{F}^{k}_v \in \mathbb{R}^{3 \times S \times S \times S}$. These are arranged into UV space as in~\cite{remelli2022dva,lombardi2021mvp} to produce volumetric texture maps of size $\{\mathbf{F}^{k}_v\}_{k=1}^{N} {\in} \mathbb{R}^{3 \times wS \times wS \times S}$, with $w{=}64$ the number of primitives per side of the UV map layout, and $N=w{\times} w$ the total number of primitives, and similarly for light directions.
\Cref{fig:overview_teacher} illustrates the overall architecture of the teacher model.

While the spatially aligned light directions are computed in a model-centric space in~\cite{bi2021deep}, this global parameterization leads to severe overfitting for articulated objects because it ignores local orientation changes produced by articulation. 
To address this, we propose to reorient view and light directions into primitive-centric coordinates.
More specifically, the view directions $\mathbf{F}^k_v$ at each primitive $k$ are represented as follows:
\begin{equation}
    \left[\mathbf{F}_{v}^{k}\right]_j = \mathbf{R}_k^{\intercal} \left(\mathbf{v} - \mathbf{p}_{k,j}\right)||\mathbf{v} - \mathbf{p}_{k,j}||_2^{-1},
\end{equation}
where $[\cdot]_j$ indexes voxels inside the primitive, $\mathbf{R}_k^{\intercal}$ is the inverse rotation matrix of the $k$-th primitive and $\mathbf{p}_{k,j}$ denotes the 3D location of the $j$-th voxel inside the $k$-th primitive.
Similarly, the light directions $\mathbf{F}_l^{k,i}$ are expressed as follows:
\begin{equation}
    \left[\mathbf{F}_{l}^{k,i}\right]_j = \mathbf{R}_k^{\intercal} \left(\mathbf{l}_i - \mathbf{p}_{k,j}\right)||\mathbf{l}_i - \mathbf{p}_{k,j}||_2^{-1},
\end{equation}
where $\mathbf{l}$ is the location of the point light. 

Additionally, joint features are input at the lowest resolution level of the U-Net layer such that the resulting appearance explicitly accounts for pose-dependent texture changes, such as small wrinkles, that may not be represented by the primitive geometry. We use a spatially aligned joint feature encoder $\mathcal{J}_t(\theta){\in}\mathbb{R}^{64\times64\times64}$ in UV space as in~\cite{bagautdinov2021driving}. Our loss is expressed by
\begin{equation}
\label{eq:loss}
    \mathcal{L} = \lambda_{MSE} \mathcal{L}_{MSE} + \lambda_{VGG} \mathcal{L}_{VGG} + \lambda_{neg} \mathcal{L}_{neg},
\end{equation}
where $\mathcal{L}_{MSE}$ is the mean-squared error between ground-truth and rendered images, $\mathcal{L}_{VGG}$ is the weighted sum of the VGG feature loss at each layer, $\mathcal{L}_{neg}$ is a regularization term and penalizes the texture with a negative intensity:

\begin{equation}
    \mathcal{L}_{neg} = \frac{\gamma}{NS^3}\sum_{k}^{N}
    ||\max(-\mathbf{C}_{k}, 0)||_2^2,
\end{equation}
with weight schedule $\gamma = \exp{\left(-\max{\left(\eta_{neg} \frac{t}{t - t_{s}}, 0\right)}\right)}$ where $t$ is the number of current iterations, and $t_s$ is the iteration when the regularization loss starts decaying.

\subsection{Student Model}
\label{sec:student}

Our student model $\mathcal{A}_{\text{env}}$ predicts the appearance of the hand model under natural illuminations represented as environment maps $\mathbf{E} \in \mathbb{R}^{M \times 3}$ as follows:
\begin{equation}
    \{\mathbf{C}_k\} = \mathcal{A}_{\text{env}}(\theta,  \mathbf{E}).
\end{equation}
The principal challenge in the student model is efficient light encoding that can be generalized to illuminations and poses unseen during training. 
The state-of-the-art model-based approach for face relighting~\cite{bi2021deep} introduces an efficient hypernet architecture conditioned on 
a $512$-dimensional bottleneck representation of a $16 \times 32$ environment
map.
We find that such a bottleneck illumination representation is difficult to generalize because it loses the spatial geometry of illumination by collapsing it into a single vector.
As we demonstrate in our evaluation (Sec.~\ref{sec:eval_student}), this issue is even more pronounced for relightable hands, as it is nontrivial to disentangle global light transport between such bottleneck illumination and pose features.
A similar observation has been made for portrait relighting, where methods based on image-aligned light features~\cite{Pandey:2021,yeh2022learning} outperform an approach using a bottleneck representation~\cite{sun2019single}. 
Inspired by this success, we propose a spatially aligned illumination representation tailored for model-based hand relighting that accurately accounts for self-occlusion due to pose changes. 

\begin{figure}
    \centering
    \setlength\rotheadsize{1.55cm}
    \setlength\tabcolsep{1pt}
    \hspace{-0.4cm}
    \begin{tabularx}{\linewidth}{l XXXX }
    \rothead{w/ Visibility} &   \includegraphics[trim={10cm 25cm 60cm 18cm},clip,width=\hsize,valign=m]{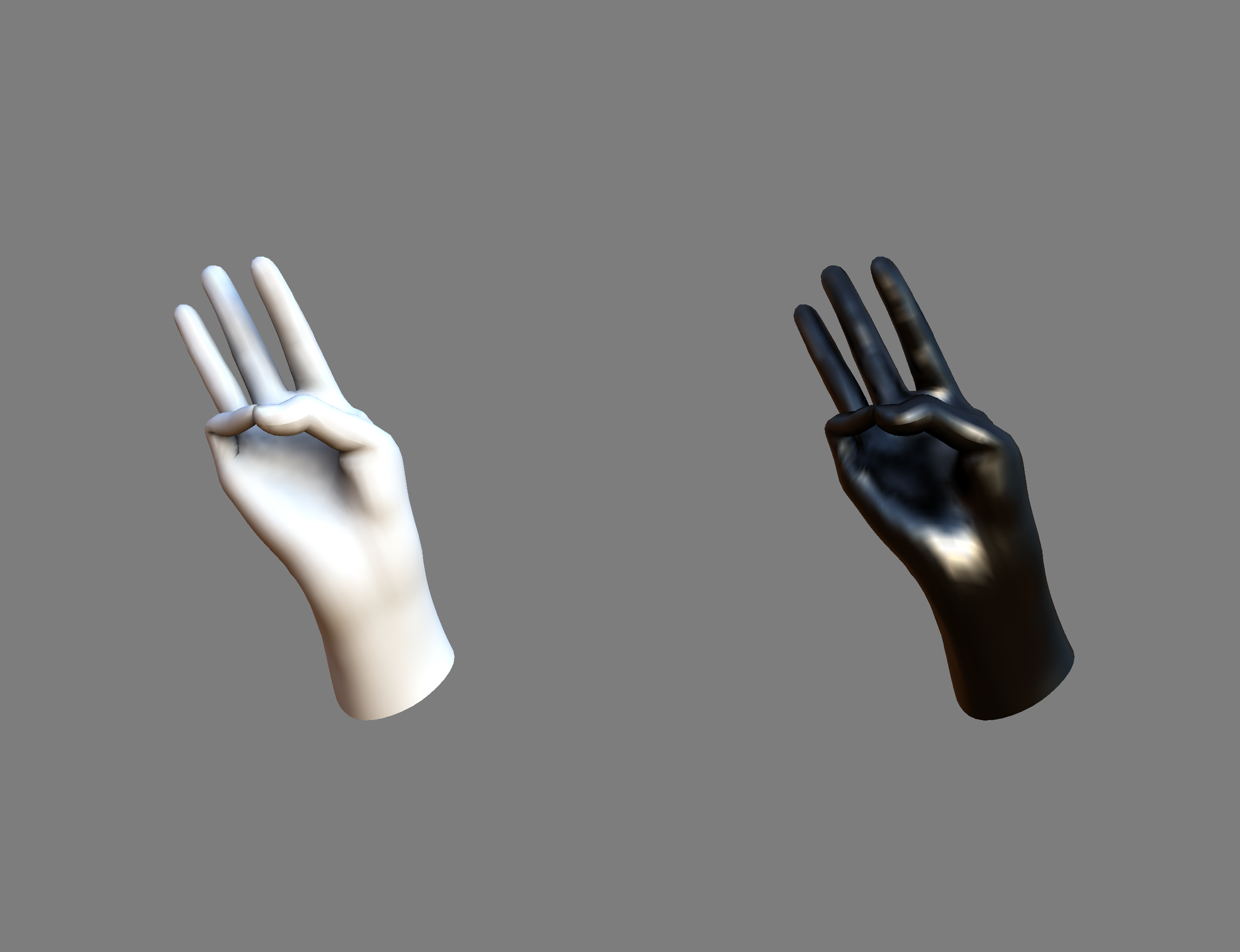}
                 &   \includegraphics[trim={56cm 25cm 14cm 18cm},clip,width=\hsize,valign=m]{figures/diffuse_specular_features/000244_features.png}
                 &   \includegraphics[trim={0cm 22cm 63cm 13cm},clip,width=\hsize,valign=m]{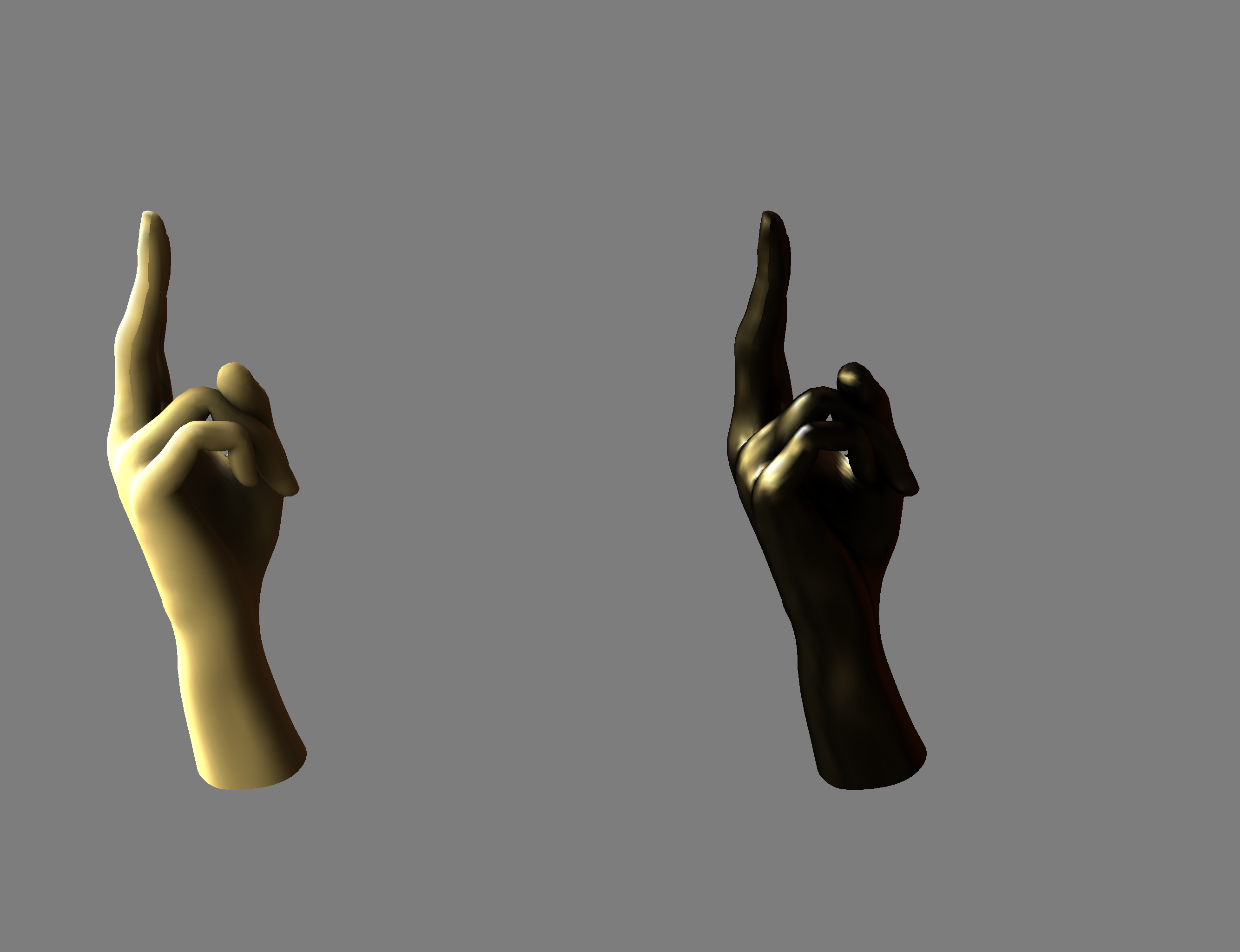}
                 &    \includegraphics[trim={45cm 22cm 18cm 13cm},clip,width=\hsize,valign=m]{figures/diffuse_specular_features/001870_features.png}     \\  \addlinespace[2pt]
    \rothead{w/o Visibility} &  \includegraphics[trim={10cm 25cm 60cm 18cm},clip,width=\hsize,valign=m]{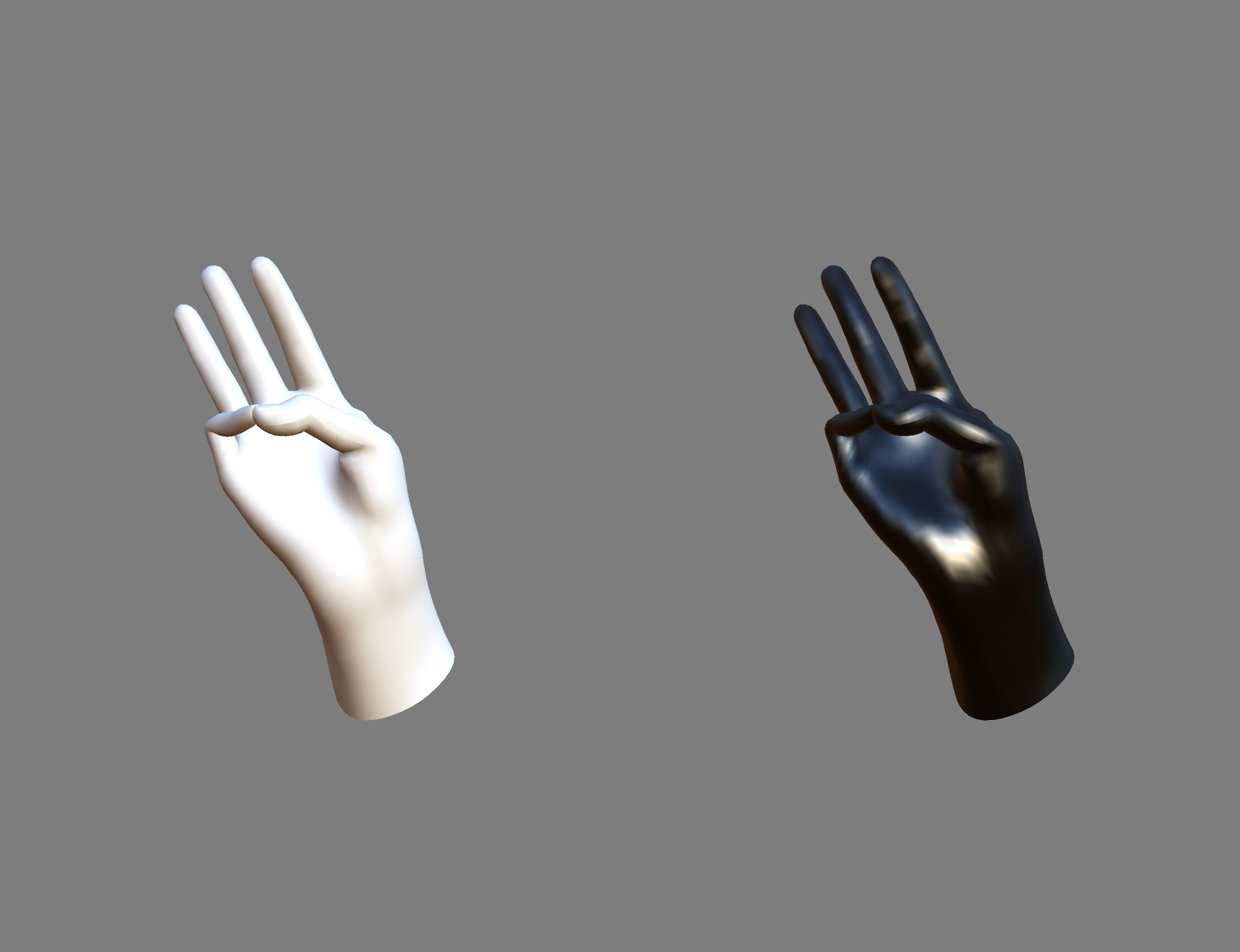} \vspace{-0.3cm} \caption*{Diffuse}
                   &   \includegraphics[trim={56cm 25cm 14cm 18cm},clip,width=\hsize,valign=m]{figures/diffuse_specular_features/000244_features_wo_shadow.png} \vspace{-0.3cm}\caption*{Specular} 
                   &   \includegraphics[trim={0cm 22cm 63cm 13cm},clip,width=\hsize,valign=m]{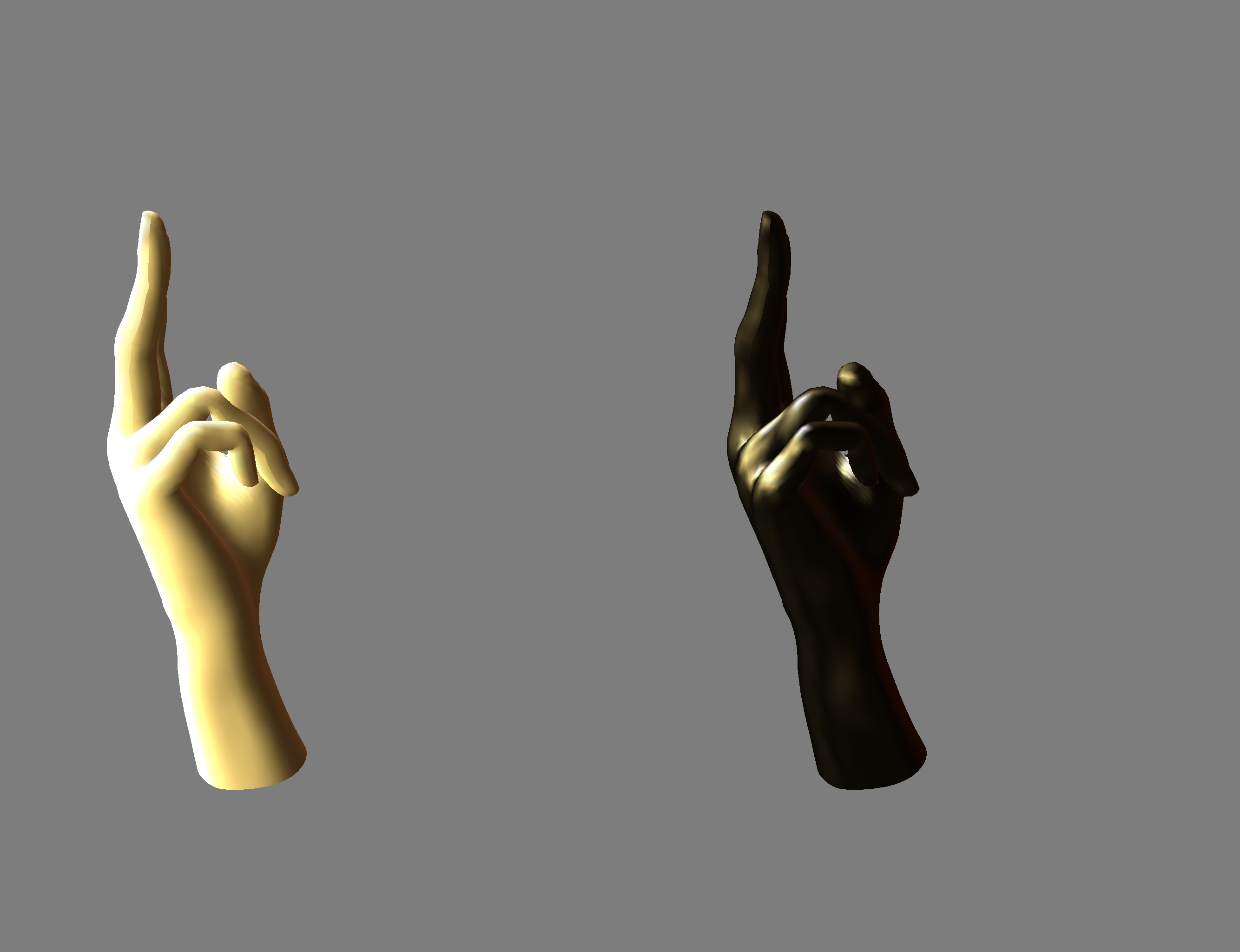} \vspace{-0.3cm} \caption*{Diffuse}
                   &    \includegraphics[trim={45cm 22cm 18cm 13cm},clip,width=\hsize,valign=m]{figures/diffuse_specular_features/001870_features_wo_shadow.png} \vspace{-0.3cm}\caption*{Specular}
    \end{tabularx}
    \vspace{-0.7cm}
    \caption{\textbf{Effect of visibility integration.} The visibility integration based on ray tracing leads to accurate encoding of shadow information in both specular and diffuse features. 
    }
    \label{fig:feature_visualization}
    \vspace{-0.5cm}
\end{figure}

Namely, we produce our texel-aligned feature representation by casting $M$ rays (one per envmap location) from each vertex and compute a weighted sum of envmap values to produce diffuse and specular components.
We incorporate the visibility
information by setting the contribution of rays hitting other mesh parts to zero.
\Cref{fig:feature_visualization} shows the effect of the visibility integration.
In practice, we use a coarse mesh to compute per-vertex features, which are then projected to texel-aligned space via barycentric interpolation. 
The projected features are fed into a fully convolutional decoder, retaining the spatial alignment between features and the output appearance $\{\mathbf{C}_k\}$.
\Cref{fig:overview_student} illustrates the overall architecture of the student model.
 
More precisely, the diffuse feature $\mathbf{d}_i\in\mathbb{R}^3$ at vertex $i$ is represented with Lambertian BRDF computed as follows:
\begin{equation}
\mathbf{d}_i = \sum_{m=1}^M \mathbf{E}(\mathbf{r}^m_i) h_i(\mathbf{r}^m_i) \max(\mathbf{n}_i \cdot \mathbf{r}^m_i, 0) \; ,
\label{eq:student_diffuse_feature}
\end{equation}
where $\mathbf{E}(\mathbf{r}^m_i) \in \mathbb{R}^3$ is the vector of the envmap intensity sampled along the ray direction $\mathbf{r}^m_i \in \mathbb{R}^3$ (based on a far-field environment assumption), 
$h_i(\mathbf{r}^m_i)$ is a binary visibility term, and $\mathbf{n}_i \in \mathbb{R}^3$ is a vertex normal.
The specular feature $\mathbf{s}_i(\alpha)\in\mathbb{R}^{3}$ is represented with Phong specular BRDF computed as 
\begin{equation}
\mathbf{s}_i(\alpha) = \sum_{m=1}^M \mathbf{E}(\mathbf{r}^m_i) h_i(\mathbf{r}^m_i) \max(\hat{\mathbf{v}}_i \cdot \mathbf{r}^m_i, 0)^{\alpha} \; ,
\label{eq:student_specular_feature}
\end{equation}
where $\alpha$ is a shininess coefficient, and $\hat{\mathbf{v}}_i$ is the view direction reflected around the normal.
To account for spatially varying material properties on hands, we take specular features with multiple shininess values ($16, 32, 64$). The feature maps are also concatenated with spatially aligned joint features $\mathcal{J}_s(\theta){\in}\mathbb{R}^{64\times128\times128}$.

Note that we train the student model $\mathcal{A}_{\text{env}}$ using the same losses used for the teacher model (Eq.\ref{eq:loss}).

\subsection{Implementation Details}
To train the teacher and student models, we use Adam~\cite{kingma:adam} optimizer and set the hyperparameters $\lambda_{MSE}$, $\lambda_{VGG}$, and $\lambda_{neg}$ to $1.0$, $1.0$, and $0.01$ respectively. We train each of the geometry module, teacher model, and student model for $100,000$ iterations with the learning rate of $0.001$, and batch size of $4$, $2$, and $4$, respectively, on NVIDIA V100 and A100. In addition, we use $N {=} 4096$ primitives whose per-axis resolution $S$ is $16$. We describe the detailed network architecture in the supplemental. To reduce redundancy in our training data in terms of poses, we adopt importance sampling based on kernel density estimation using the subset of tracked hand vertices in root-normalized coordinates. We generate $25,000$ images with $1000$ frames and $25$ cameras to train the student model. To compute the texel-aligned lighting features, we use envmap of size $M {=} 512 (16 \times 32)$, and coarse mesh with $2825$ vertices. The visibility is computed with a GPU-accelerated triangle-ray intersection using NVIDIA OptiX~\cite{parker2010optix}.

\begin{figure}[t]
    \setlength\tabcolsep{1pt}
    \hspace{-0.4cm}
    \begin{tabularx}{\linewidth}{l XXX }
    \rothead{GT} &   \includegraphics[trim={0cm 17cm 10cm 18cm},clip,width=\hsize,valign=m]{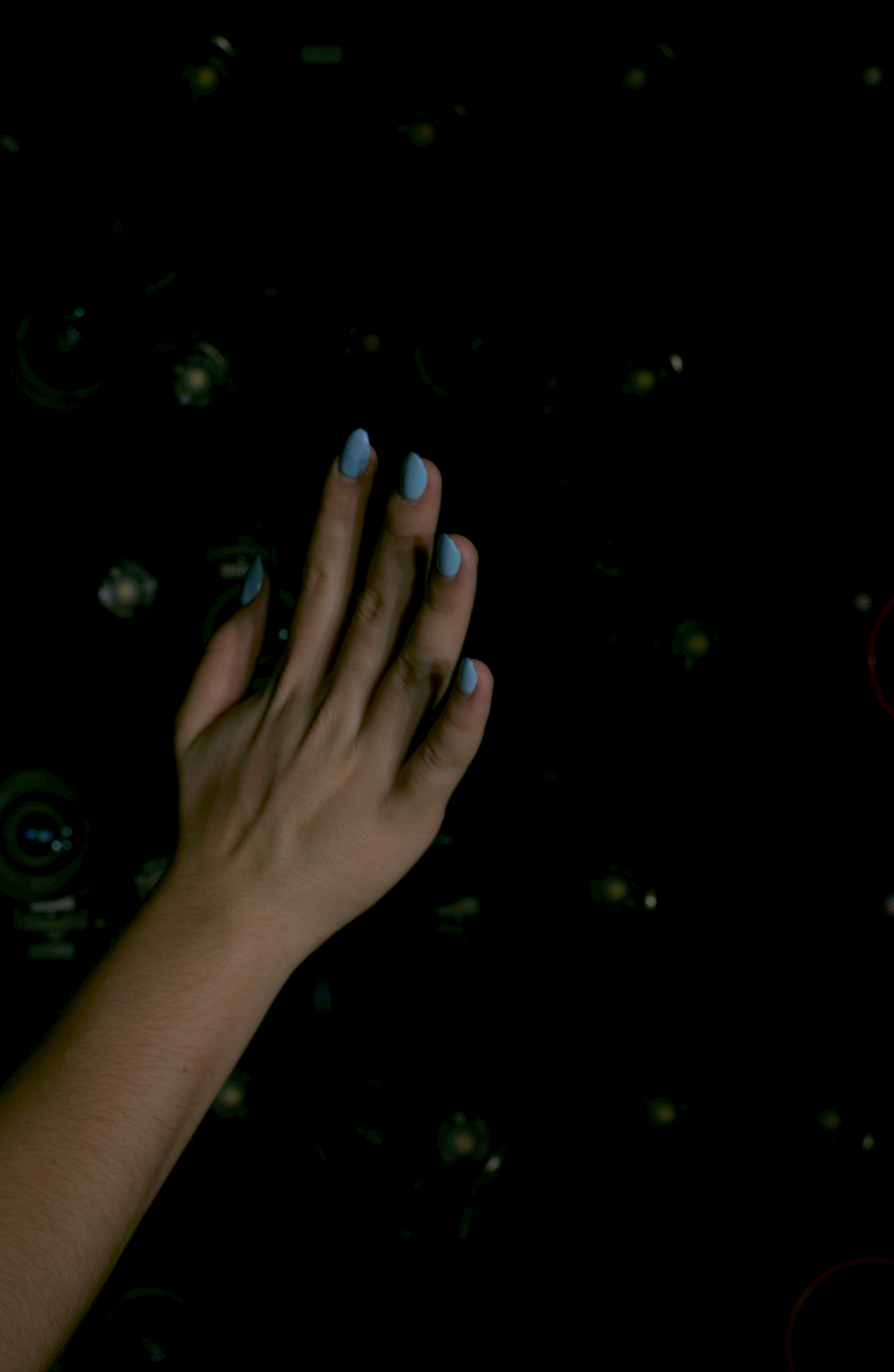}
                 &   \includegraphics[trim={10cm 25cm 10cm 20cm},clip,width=\hsize,valign=m]{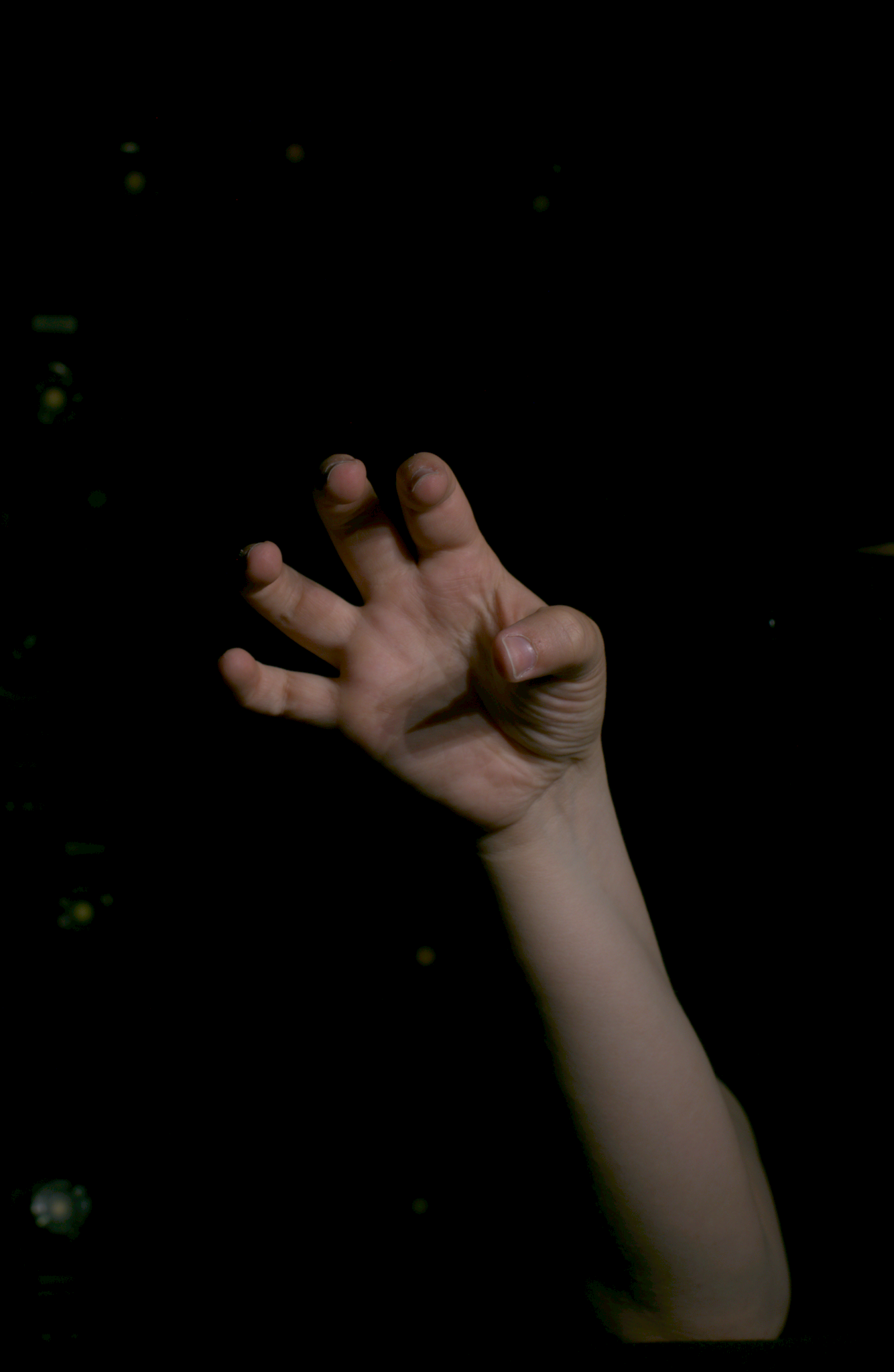}
                 &   \includegraphics[trim={15cm 28cm 5cm 17cm},clip,width=\hsize,valign=m]{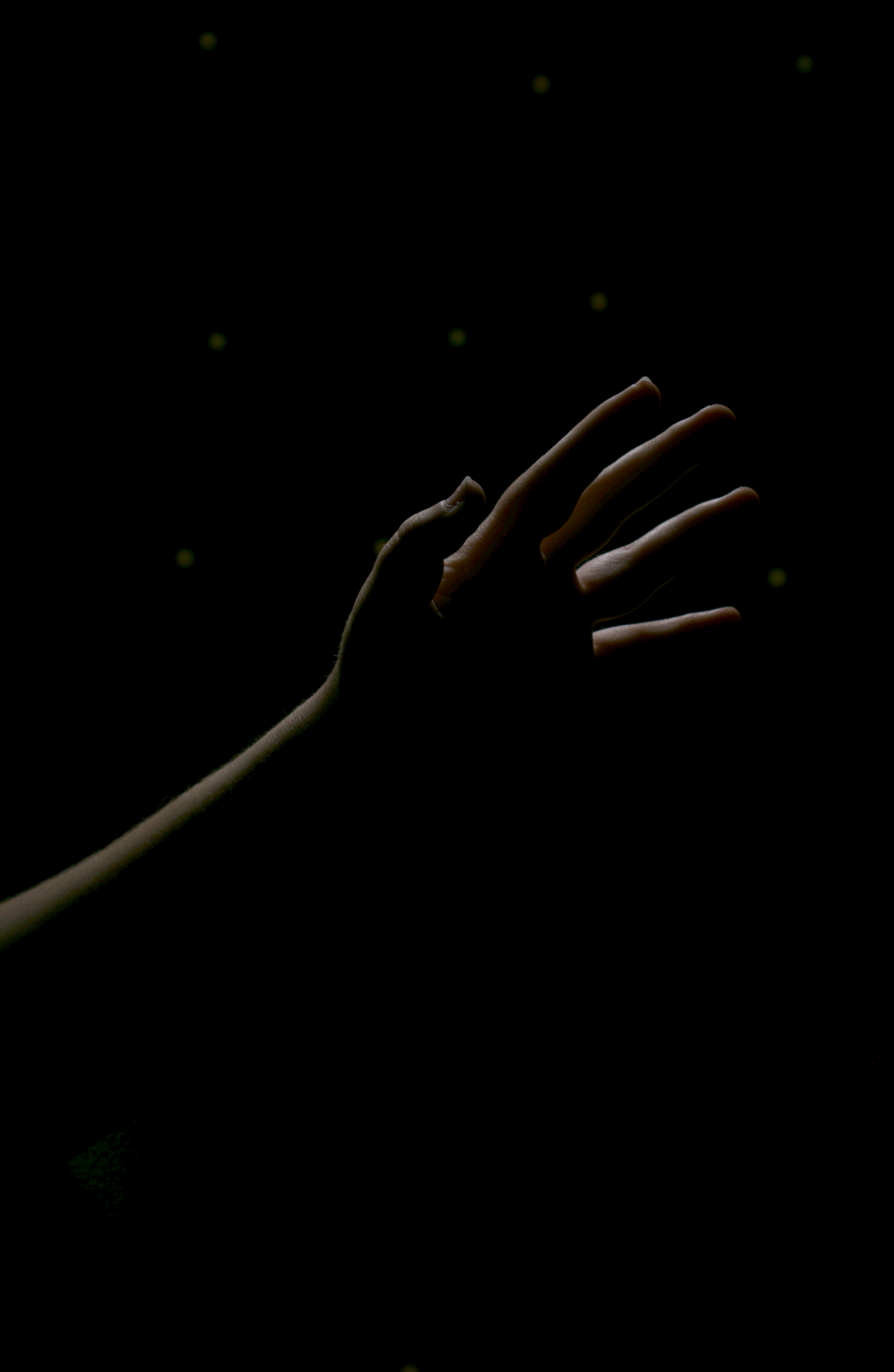}     \\  \addlinespace[2pt]
    \rothead{Ours} &  \includegraphics[trim={0cm 17cm 10cm 18cm},clip, width=\hsize,valign=m]{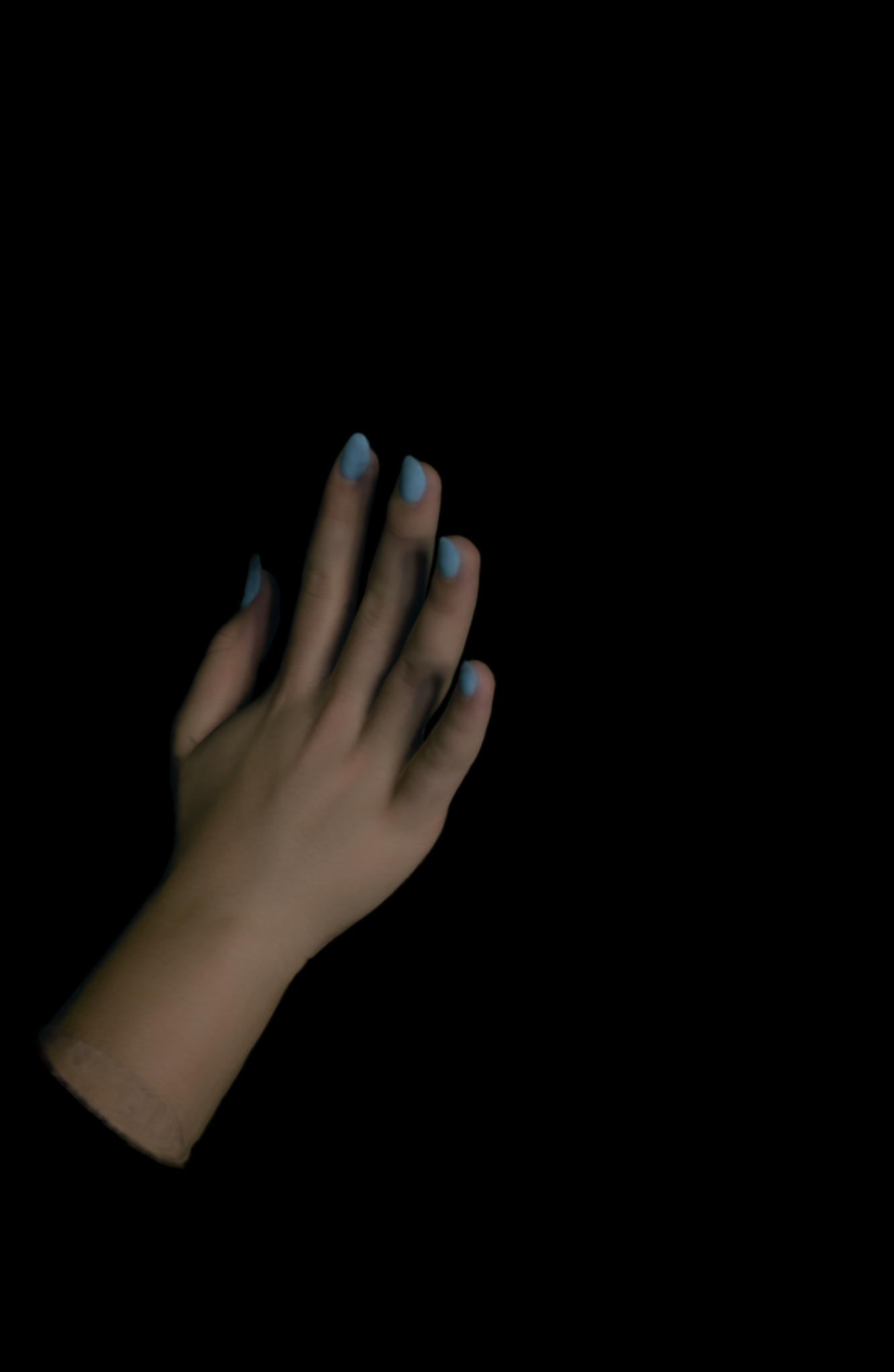}
                   &   \includegraphics[trim={10cm 25cm 10cm 20cm},clip,width=\hsize,valign=m]{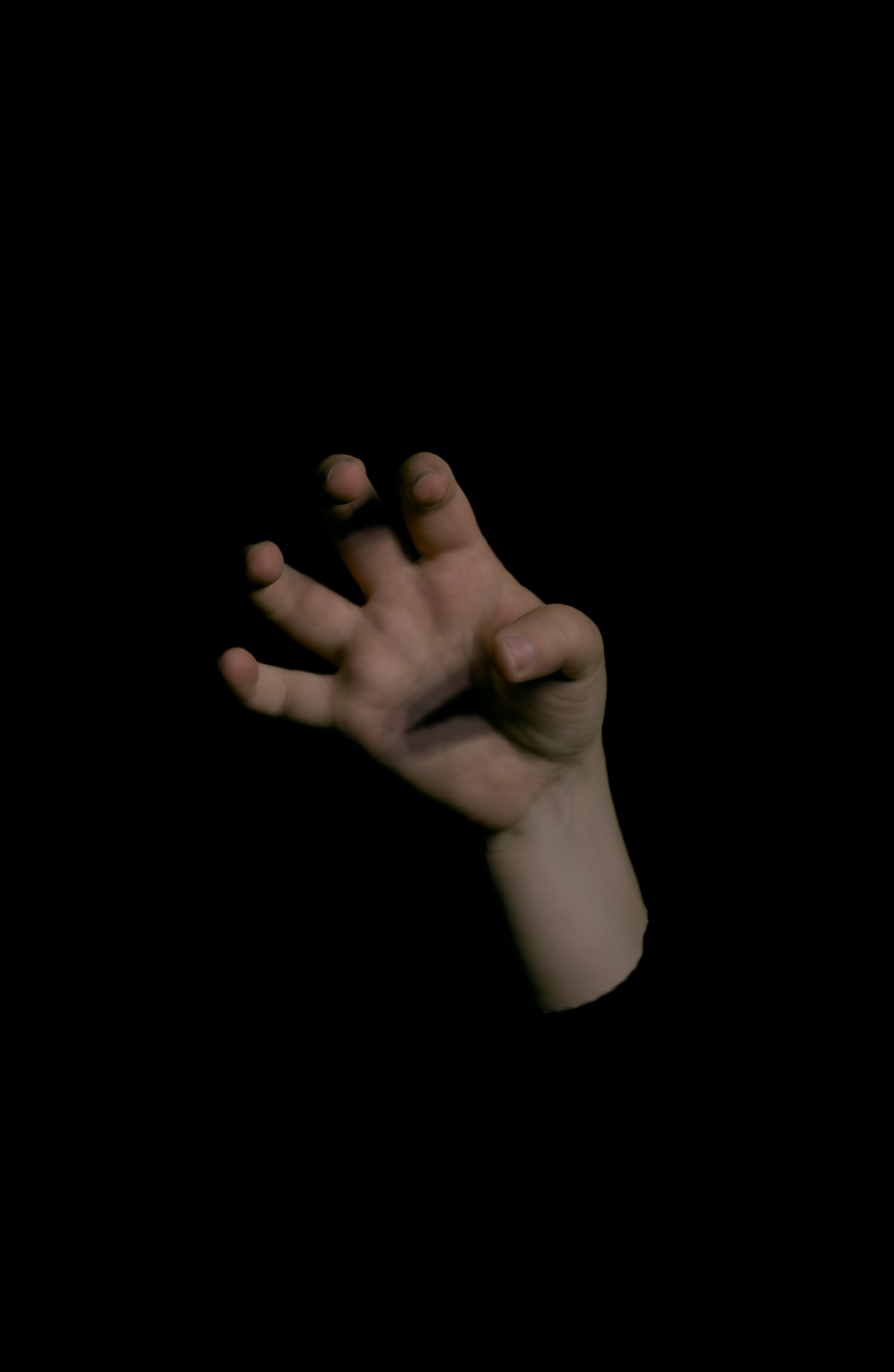}
                   &   \includegraphics[trim={15cm 28cm 5cm 17cm},clip,width=\hsize,valign=m]{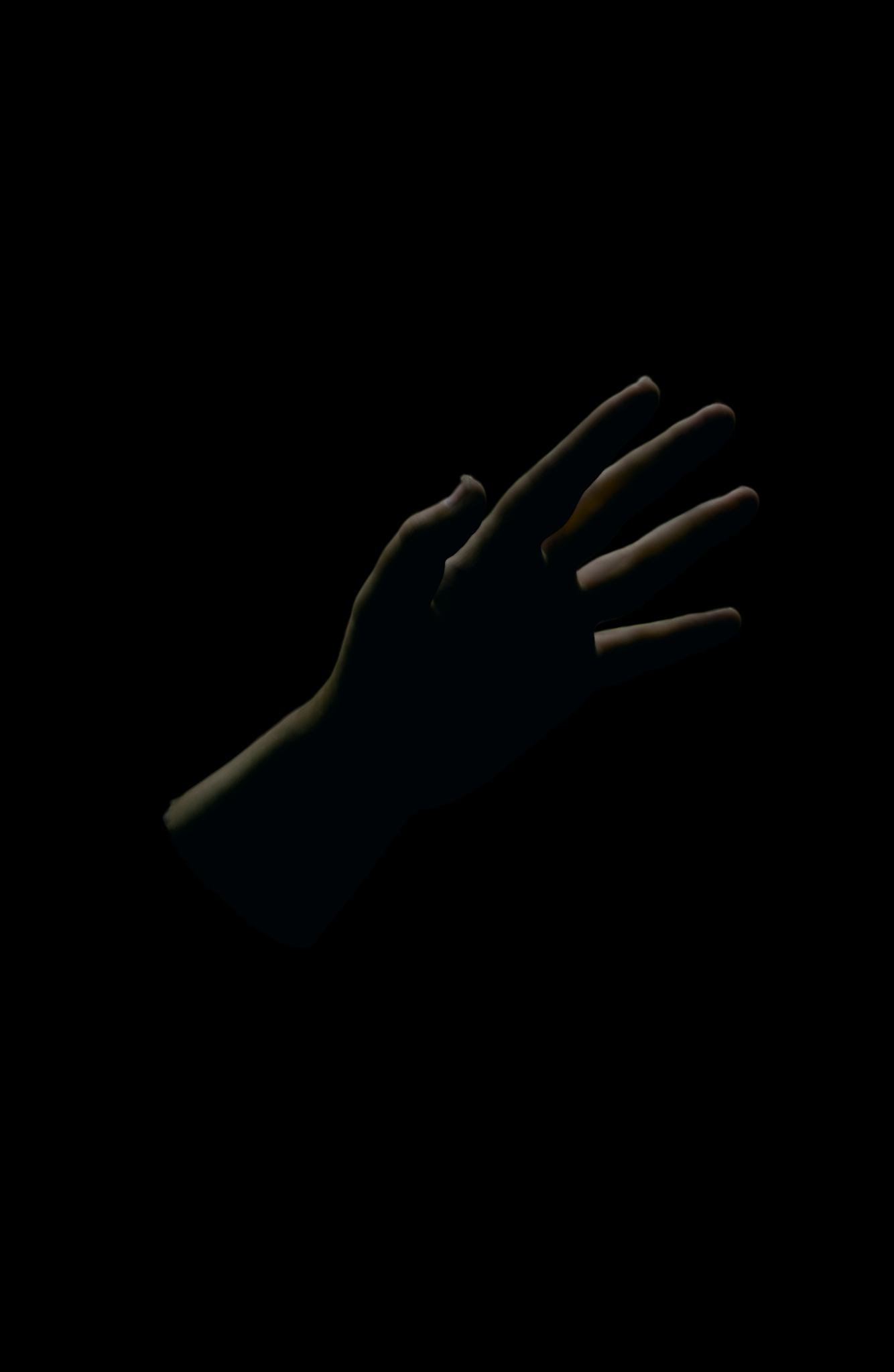}
    \end{tabularx}
    \caption{\textbf{Qualitative results of the teacher model.} We evaluate our method using the ground-truth images captured in the light-stage. Our approach successfully models inter-reflection, shadow, and subsurface scattering.}
    \label{fig:teacher_qualitative_evaluation}
    \vspace{-0.5cm}
\end{figure}

\section{Experimental Results}
We evaluate our method using 2 subjects with left, right and two hands captured in a light-stage as described in Sec.~\ref{sec:pre}. For evaluation, we exclude several segments to assess the generalization of our model to novel poses. To evaluate the generalization of the student model to novel illumination, we use 3094 high-resolution HDR environment maps consisting of the ones in \cite{gardner2017learning} and \cite{sun2019single}. We use 2560 of them for training and 534 for evaluation. Note that the two-hand sequences are used only for teacher model evaluation. We also compare our approach with the state-of-the-art model-based relighting method~\cite{bi2021deep}. As \cite{bi2021deep} was originally proposed for face relighting, we make several modifications for fair comparison. Please refer to the supplemental for details. 

We report mean squared error (MSE) and Structural Similarity Index (SSIM) to measure the quality of the generated images by the teacher and student models.
To solely evaluate the quality of hands, we remove the background by using a mask image obtained from the tracked hand geometry.

\subsection{Evaluation of Teacher Models}

We first evaluate the effectiveness of the proposed teacher model using the images captured with the light-stage as ground-truth.

\begin{table*}[t]
    \centering
    \setlength\extrarowheight{1pt}
\small{
    \scalebox{0.85}{
    \begin{tabular}{c||ccc|ccc|ccc|ccc}
        & \multicolumn{6}{c|}{Subject 1} & \multicolumn{6}{c}{Subject 2} \\ \hline
        & \multicolumn{3}{c|}{MSE ($\times 10^{-3}$) $\downarrow$} & \multicolumn{3}{c|}{SSIM $\uparrow$} & \multicolumn{3}{c|}{MSE ($\times 10^{-3}$) $\downarrow$} & \multicolumn{3}{c}{SSIM $\uparrow$} \\ \hline
        & Right & Left & Both & Right & Left & Both & Right & Left & Both & Right & Left & Both \\ \hline 
        Ours                                   & \textbf{4.9126} & \textbf{5.8608} & \textbf{15.7589} & \textbf{0.9790} & \textbf{0.9805} & \textbf{0.9536} & \textbf{8.8205} & \textbf{7.9357} & \textbf{22.3559} & \textbf{0.9541} & \textbf{0.9559} & \textbf{0.9075} \\
        w/o Visibility                             & 7.3201 & 7.8870 & 22.8308 & 0.9773 & 0.9792 & 0.9488 & 9.7104 & 9.9781 & 26.5647 & 0.9536 & 0.9543 & 0.9050 \\
    \end{tabular} 
    }
}
    \vspace{-0.2cm} 
    \caption{\textbf{Quantitative comparison of the teacher model.} We measure the MSE and SSIM metrics on the right, left, and two-hand sequences. The result shows that conditioning visibility significantly improves generalization to test poses and illuminations.}
    \label{tab:teacher_evaluation}
    \vspace{-0.3cm} 
\end{table*}

\noindent
{\bf   Qualitative Evaluation.}
We evaluate the quality of the images generated by our model against the real images captured in the light-stage.
As shown in~\Cref{fig:teacher_qualitative_evaluation}, our teacher model is able to reproduce diverse pose-dependent appearance under multiple point-light sources, such as shadows on the wrinkles and reflection on the skin and nails.

\begin{figure}
    \centering
    \setlength\extrarowheight{1pt}
    \begin{minipage}{0.15\textwidth}
       \centering
       \includegraphics[trim={5cm 20cm 5cm 15cm},clip,width=1.0\textwidth]{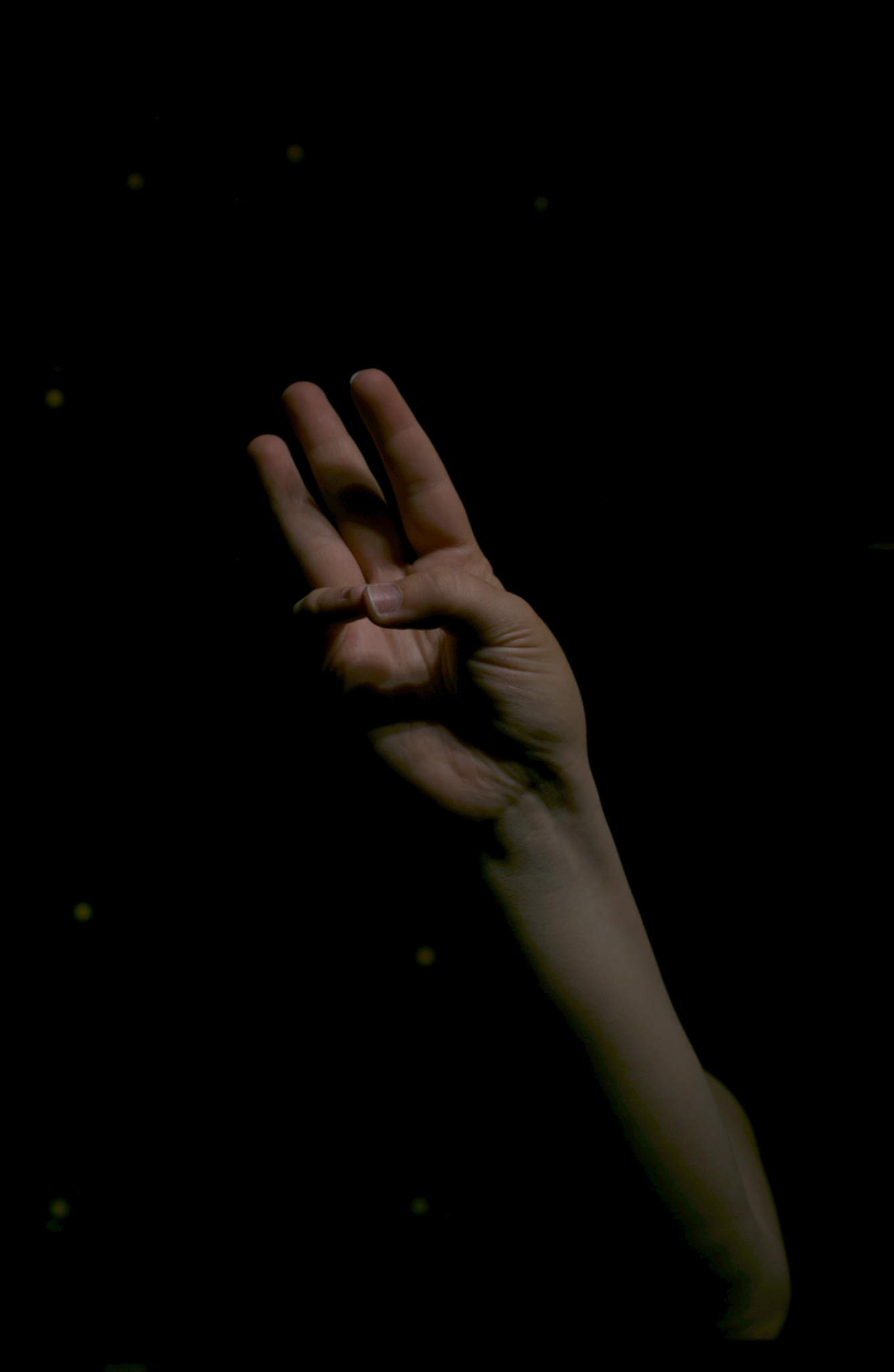}
    \end{minipage}
    \begin{minipage}{0.15\textwidth}
       \centering
       \includegraphics[trim={5cm 20cm 5cm 15cm},clip,width=1.0\textwidth]{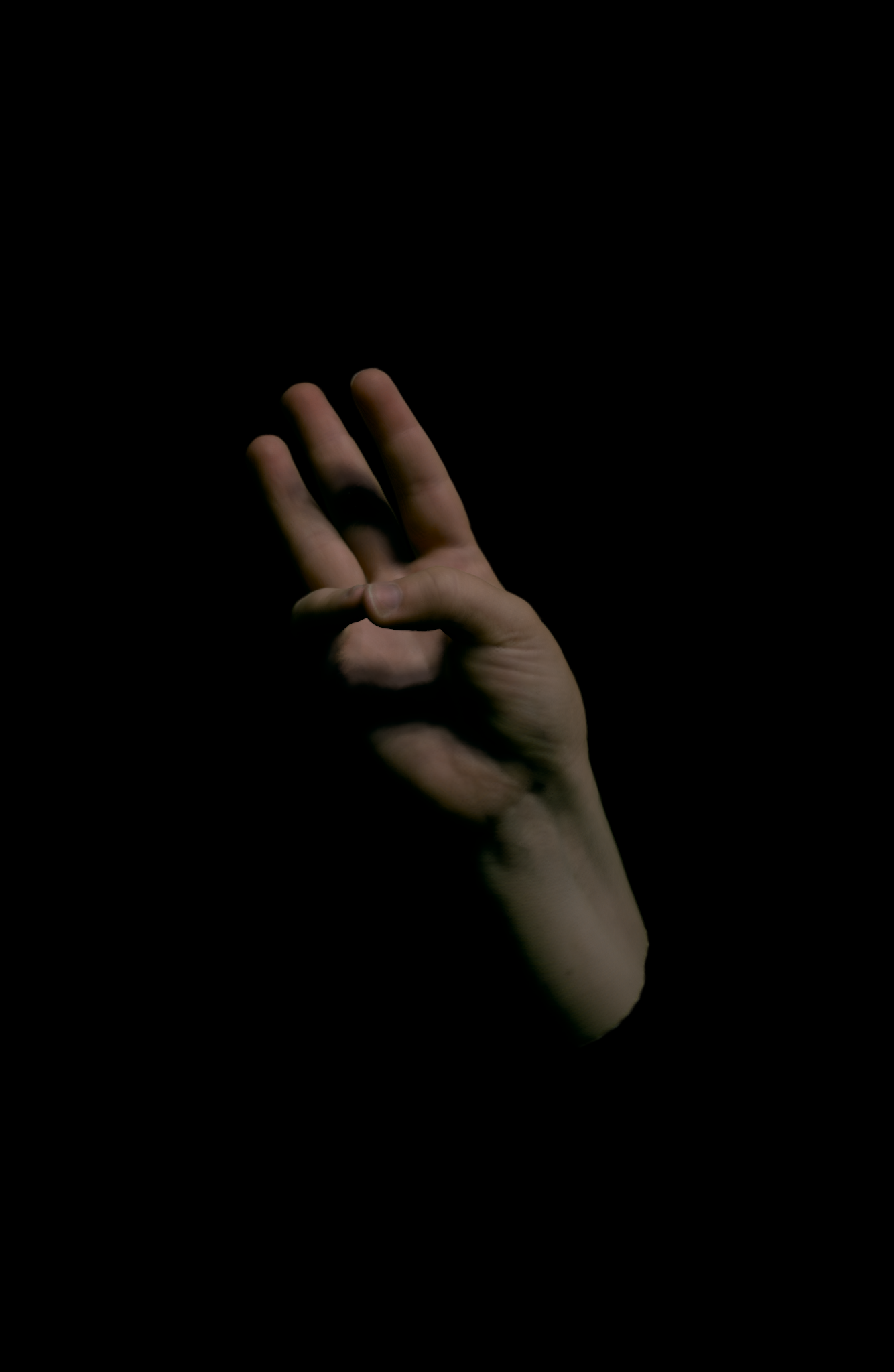}
    \end{minipage}
    \begin{minipage}{0.15\textwidth}
       \centering
       \includegraphics[trim={5cm 20cm 5cm 15cm},clip,width=1.0\textwidth]{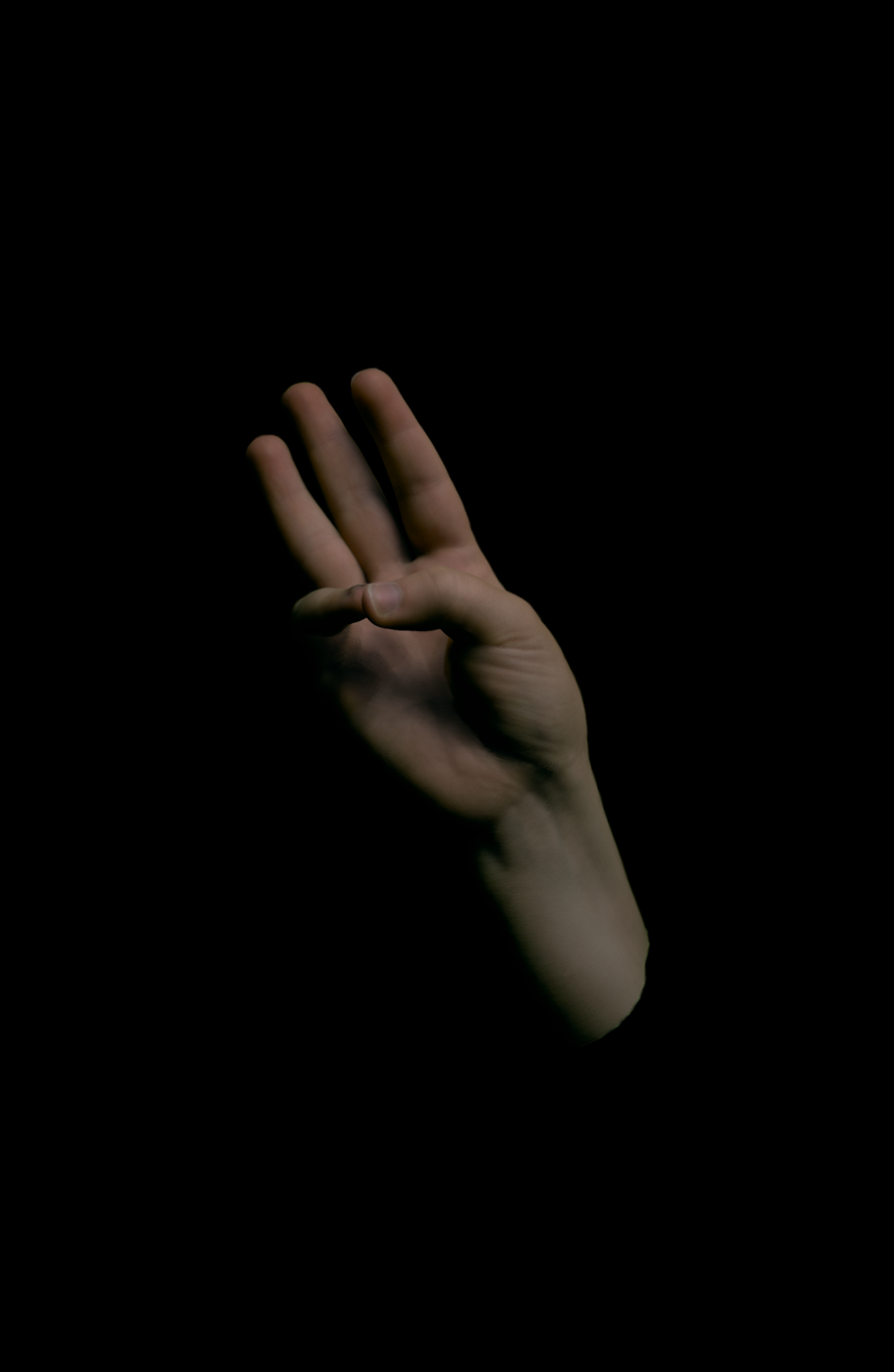}
    \end{minipage} \\
    \begin{minipage}{0.15\textwidth}
       \centering
       \includegraphics[trim={2cm 20cm 8cm 15cm},clip,width=1.0\textwidth]{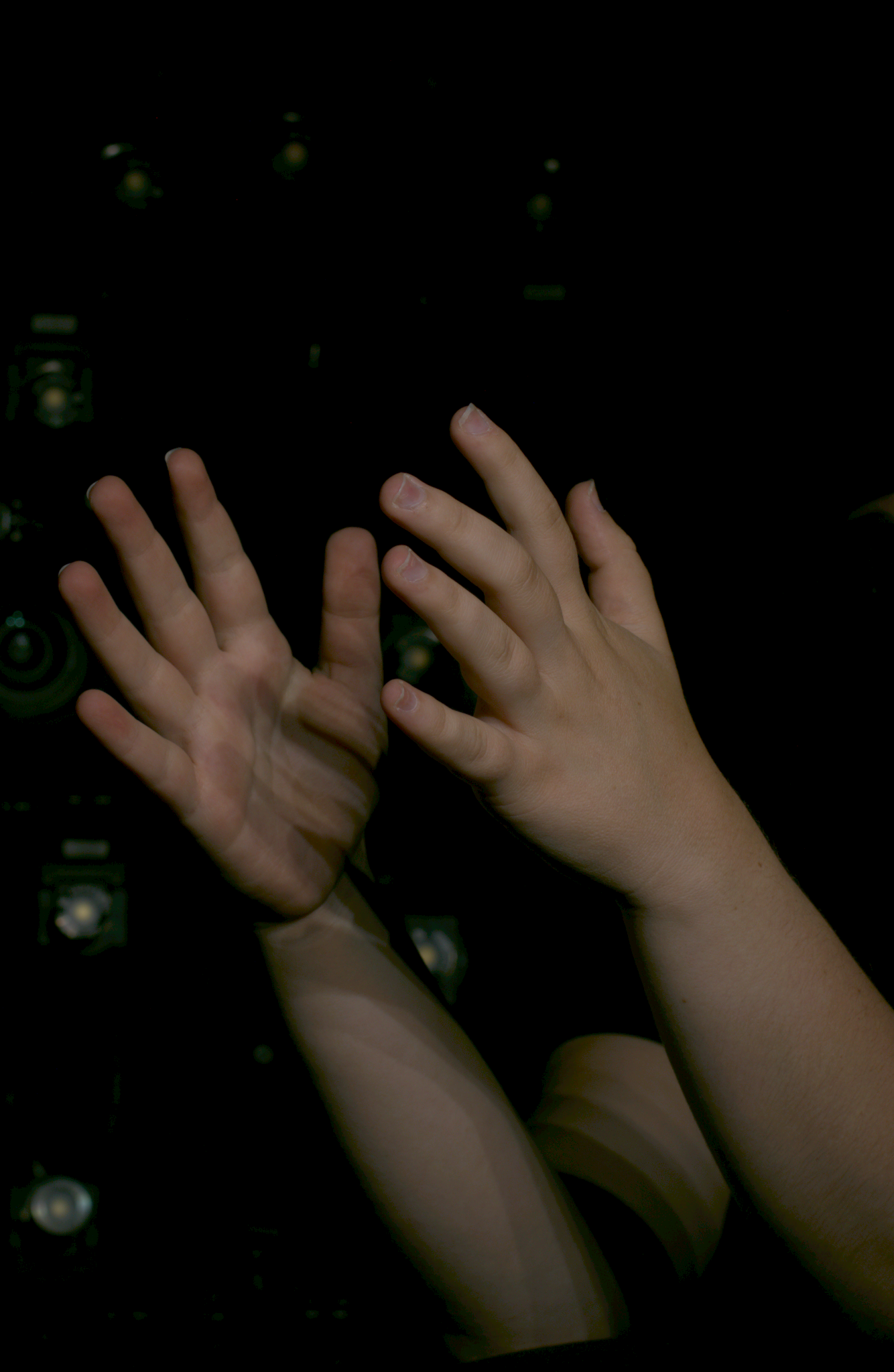}
       \caption*{Ground-Truth}
    \end{minipage}
    \begin{minipage}{0.15\textwidth}
       \centering
       \includegraphics[trim={2cm 20cm 8cm 15cm},clip,width=1.0\textwidth]{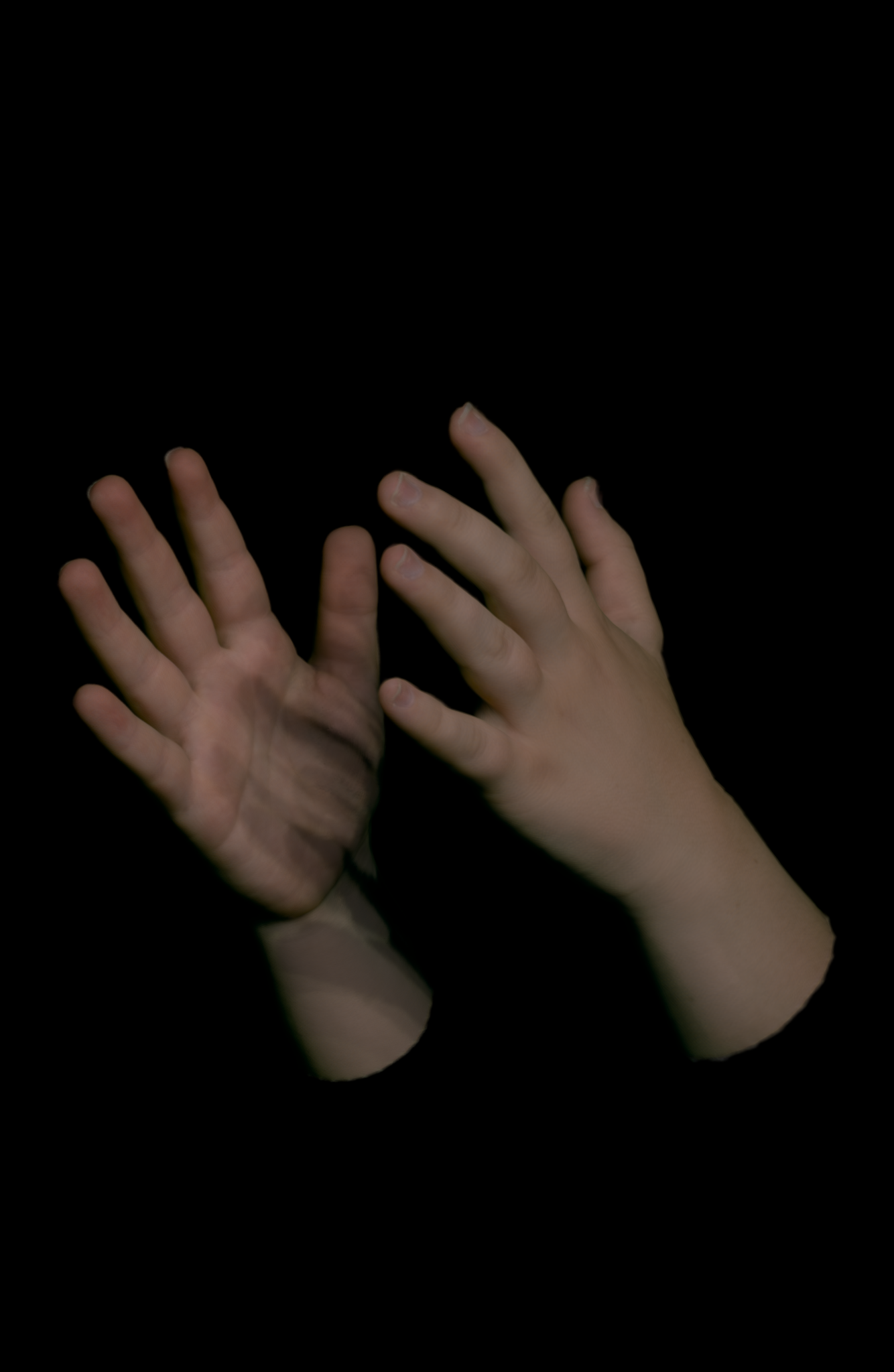}
       \caption*{w/ Visibility}
    \end{minipage}
    \begin{minipage}{0.15\textwidth}
       \centering
       \includegraphics[trim={2cm 20cm 8cm 15cm},clip,width=1.0\textwidth]{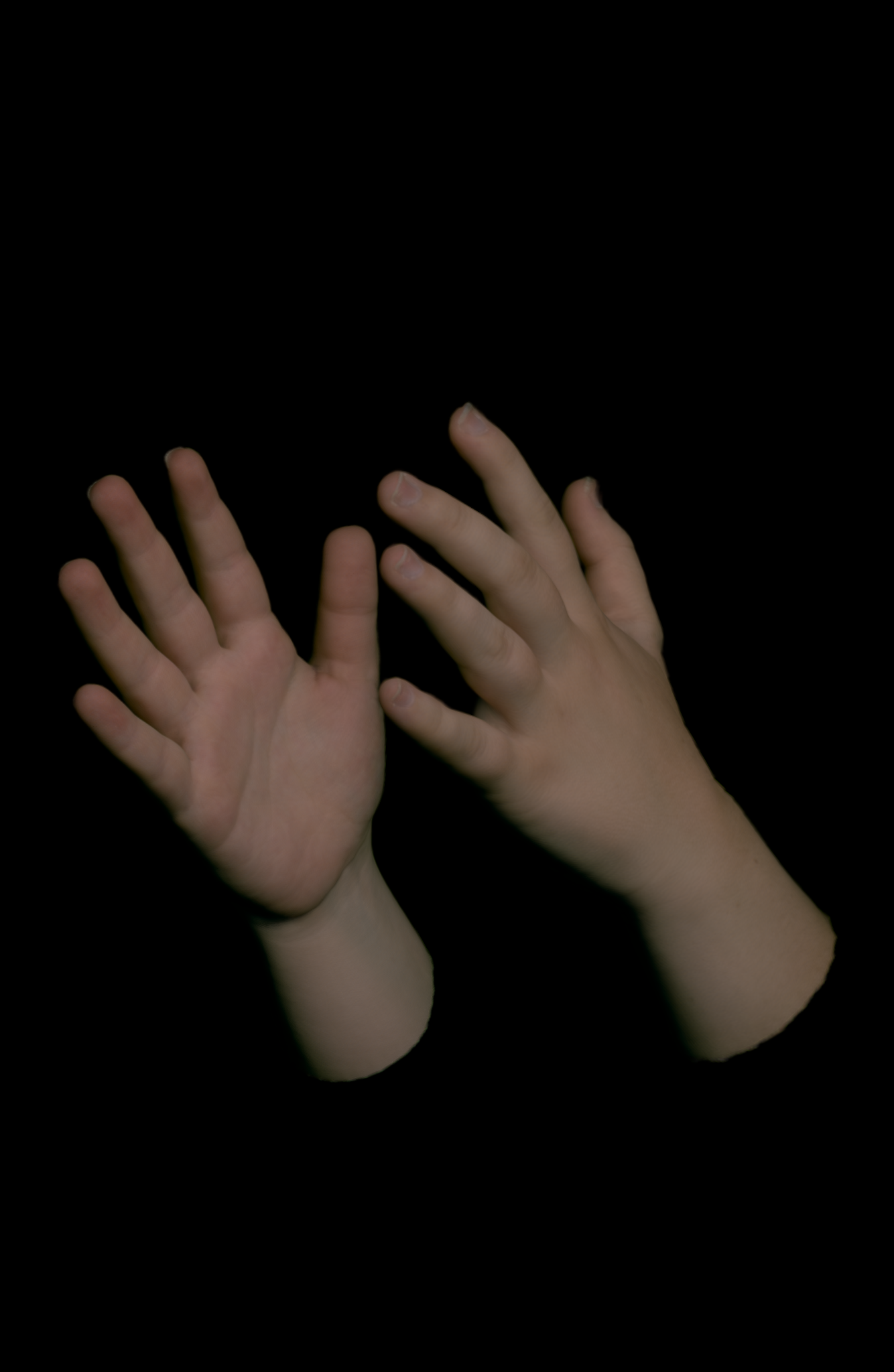}
       \caption*{w/o Visibility}
    \end{minipage}
    \caption{\textbf{Ablation on visibility conditioning of the teacher model.} The lack of visibility features leads to incorrect shadows.}
    \label{fig:comparison_shadow}
    \vspace{-0.5cm}
\end{figure}

\noindent
{\bf   Ablation on Visibility Conditioning.}
\Cref{tab:teacher_evaluation} shows that the visibility input significantly improves the accuracy on all the metrics especially for the two hand sequences. 
As shown in \Cref{fig:comparison_shadow}, while the model without visibility input can overfit to training poses by relying on the joint information, it does not generalize to unseen poses or two hand cases. Thus, the visibility conditioning is essential to generalize beyond training pose and light distributions.

\subsection{Evaluation of Student Models}
\label{sec:eval_student}
\Cref{tab:student_evaluation} shows that our method achieves the best MSE and SSIM scores on all the metrics under unseen hand poses and illumination settings.
Since we do not have ground-truth real images under natural illumination, we use test images generated by the teacher model for evaluation. 
To confirm the effectiveness of our method, we compare our method against the latest model-based relighting method~\cite{bi2021deep} and perform ablation study on visibility awareness and specular features.

\begin{table*}[t]
    \setlength\extrarowheight{1pt}
    \centering
    \small{
    \scalebox{0.85}{
    \begin{tabular}{c||ccc|ccc|ccc|ccc}
        & \multicolumn{6}{c|}{Subject 1} & \multicolumn{6}{c}{Subject 2} \\ \hline
        & \multicolumn{3}{c|}{MSE ($\times 10^{-3}$) $\downarrow$} & \multicolumn{3}{c|}{SSIM $\uparrow$} & \multicolumn{3}{c|}{MSE ($\times 10^{-3}$) $\downarrow$} & \multicolumn{3}{c}{SSIM $\uparrow$} \\ \hline
        & Right & Left & Both & Right & Left & Both & Right & Left & Both & Right & Left & Both \\ \hline 
        DRAM~\cite{bi2021deep} & 31.1372         & 24.4368  & 64.2035      & 0.9904          & 0.9927  &  0.9752         & 30.6582    & 24.7238  & 70.4215 & 0.9901          & 0.9898 & 0.9665 \\ \hline
        Ours                                & \textbf{5.4076} & \textbf{5.9600} & \textbf{4.3474} & \textbf{0.9961} & \textbf{0.9960} & \textbf{0.9915} & \textbf{5.7977}  & \textbf{7.2598} & \textbf{4.5196} & \textbf{0.9952}  & \textbf{0.9954} & \textbf{0.9881} \\
        w/o Specular                        & 5.7660          & 7.2631    & 5.0732       & 0.9956          & 0.9952     & 0.9914     & 7.1569           & 7.4892 & 4.9008 & 0.9948           & 0.9943 & \textbf{0.9881} \\
        w/o Visibility                      & 6.6110          & 8.1886     & 11.6771    & 0.9955          & 0.9948    & 0.9893      & 7.8589           & 8.5550 & 9.1859 & 0.9938 & 0.9938 & 0.9862 \\
    \end{tabular}
    }
    }
    \vspace{-0.2cm}
    \caption{\small{\textbf{Quantitative evaluation of the student model.} Our student model outperforms the state-of-the-art model-based relighting method~\cite{bi2021deep} by a large margin. In addition, the proposed visibility and specular features integration significantly improve the generalization to unseen poses and natural illuminations. }}
    \label{tab:student_evaluation}
\end{table*}

\begin{figure*}[h!]
    \centering
    \scalebox{0.95}{
    \begin{minipage}{0.19\textwidth}
       \centering
       \includegraphics[width=1.0\textwidth]{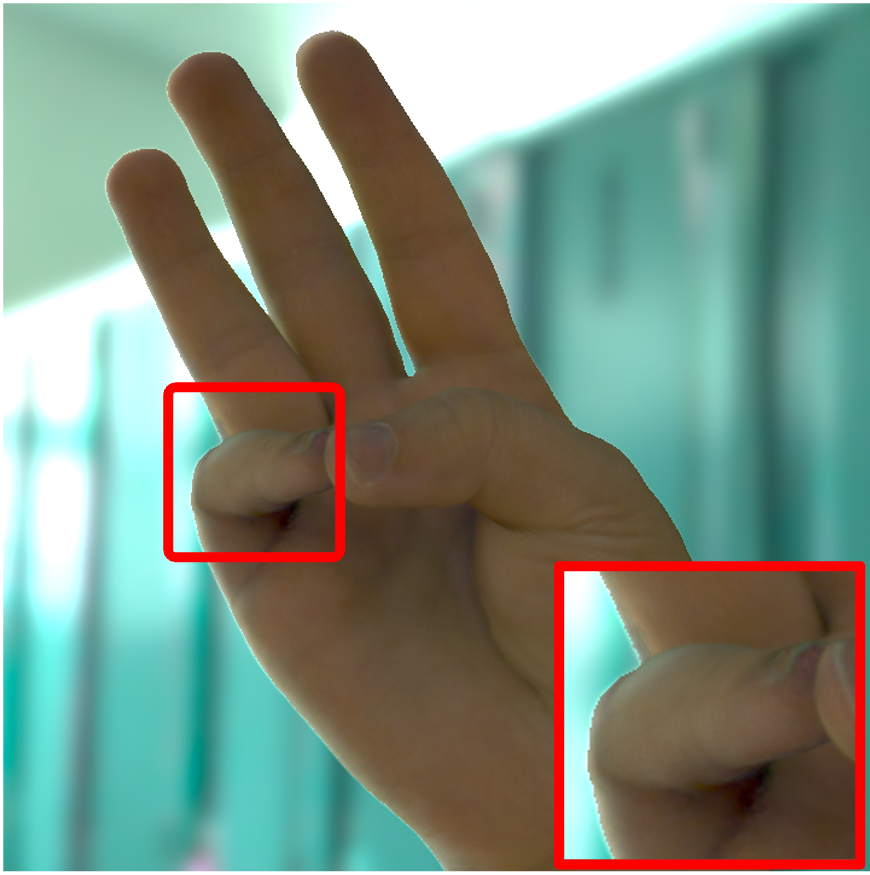}
    \end{minipage}
    \begin{minipage}{0.19\textwidth}
       \centering
       \includegraphics[width=1.0\textwidth]{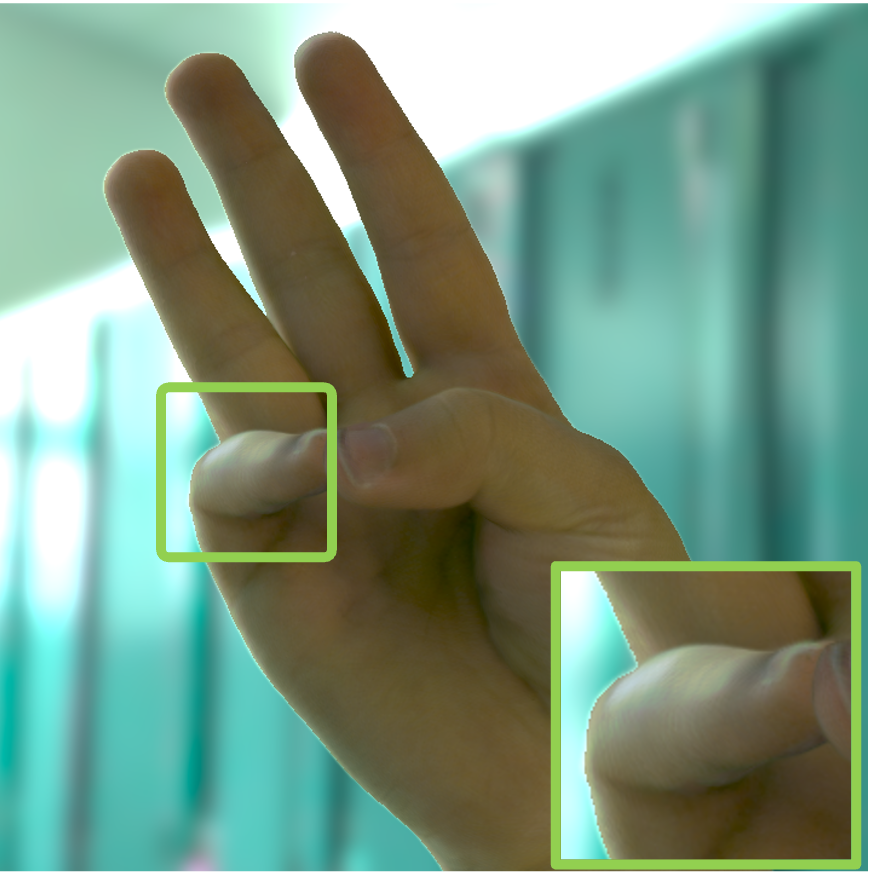}
    \end{minipage}
    \begin{minipage}{0.19\textwidth}
       \centering
       \includegraphics[width=1.0\textwidth]{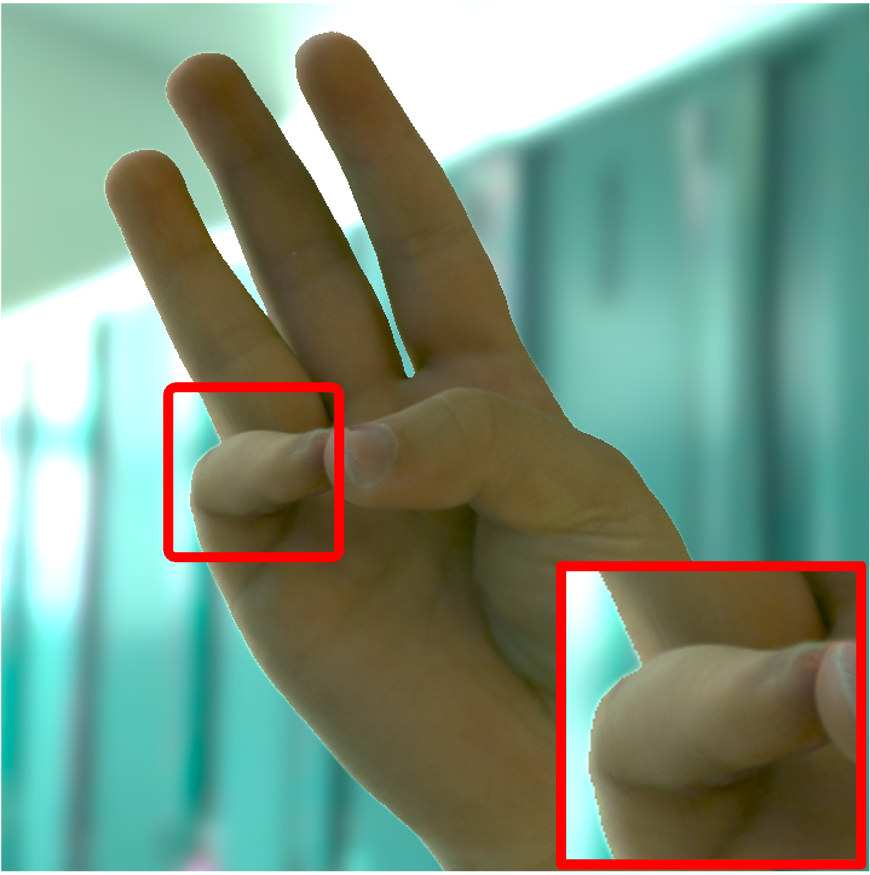}
    \end{minipage}
    \begin{minipage}{0.19\textwidth}
       \centering
       \includegraphics[width=1.0\textwidth]{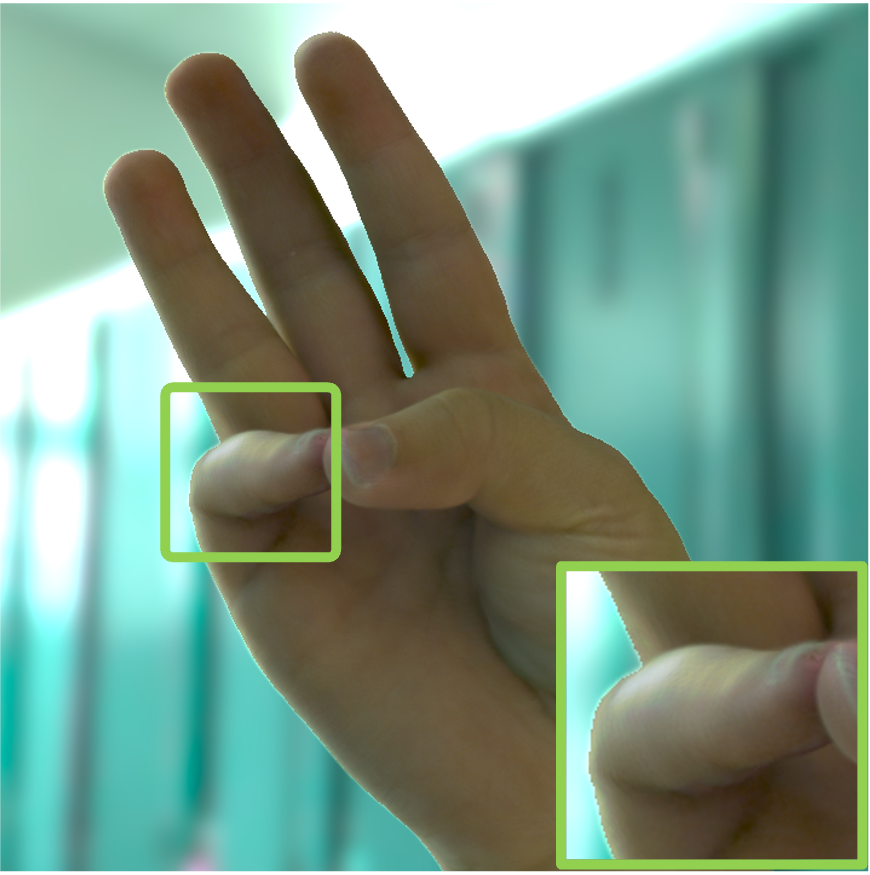}
    \end{minipage}
    \begin{minipage}{0.19\textwidth}
       \centering
       \includegraphics[width=1.0\textwidth]{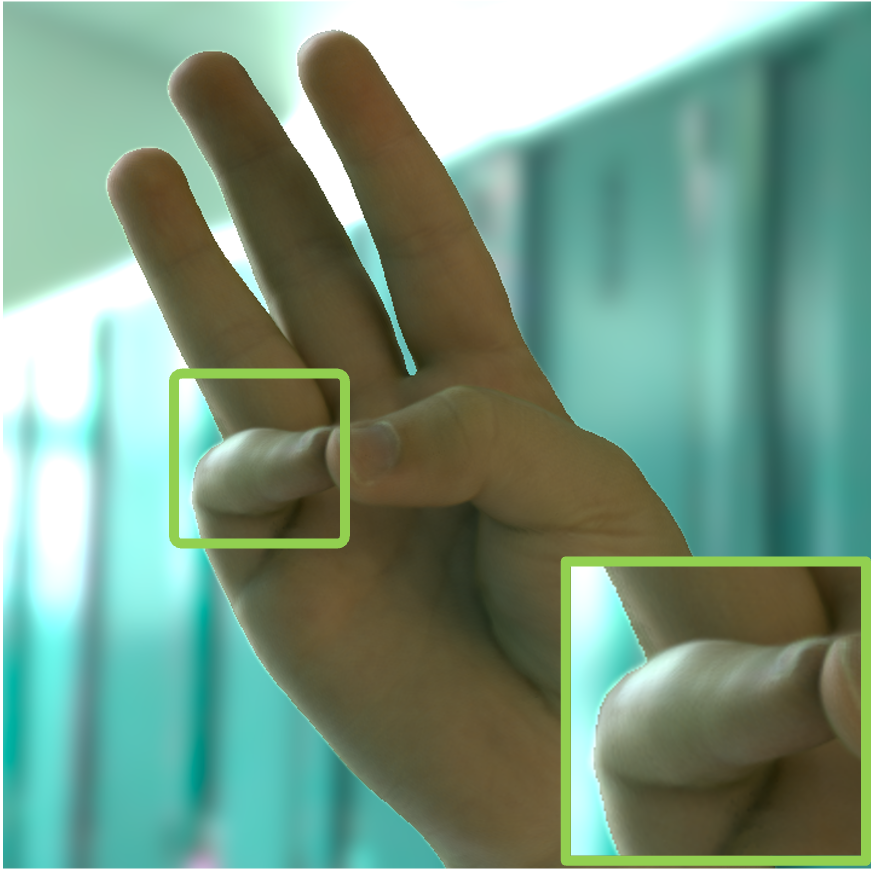}
    \end{minipage}    }
\\
    \scalebox{0.95}{

    \begin{minipage}{0.19\textwidth}
       \centering
       \includegraphics[width=1.0\textwidth]{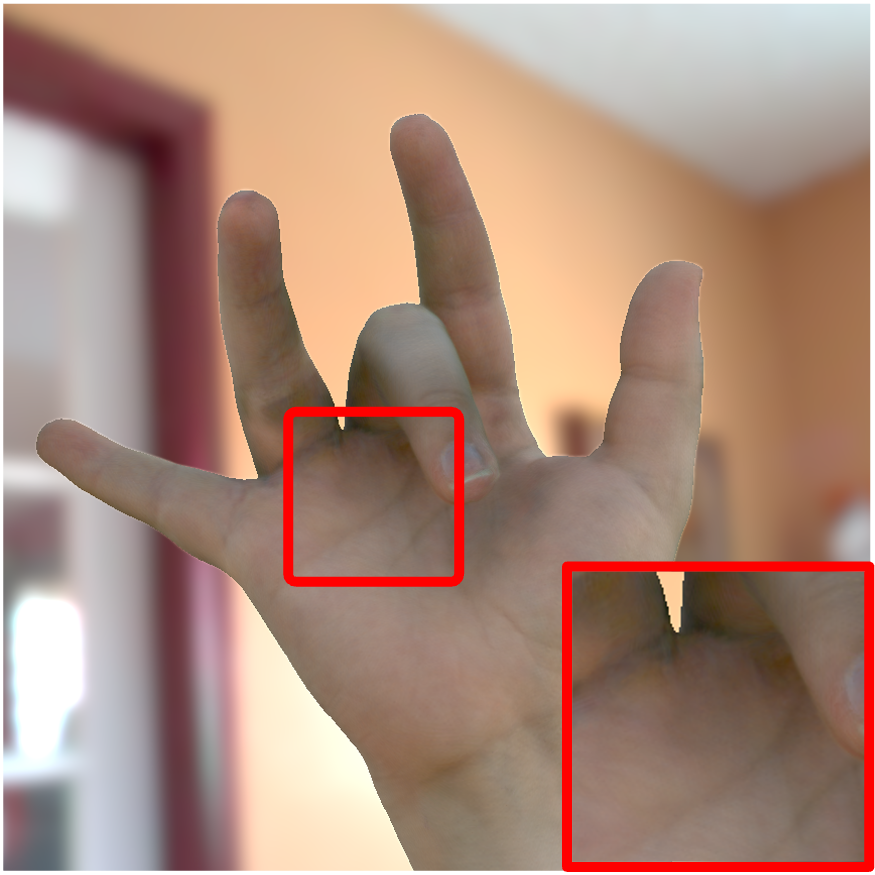}
       \caption*{DRAM~\cite{bi2021deep}}
    \end{minipage}
    \begin{minipage}{0.19\textwidth}
       \centering
       \includegraphics[width=1.0\textwidth]{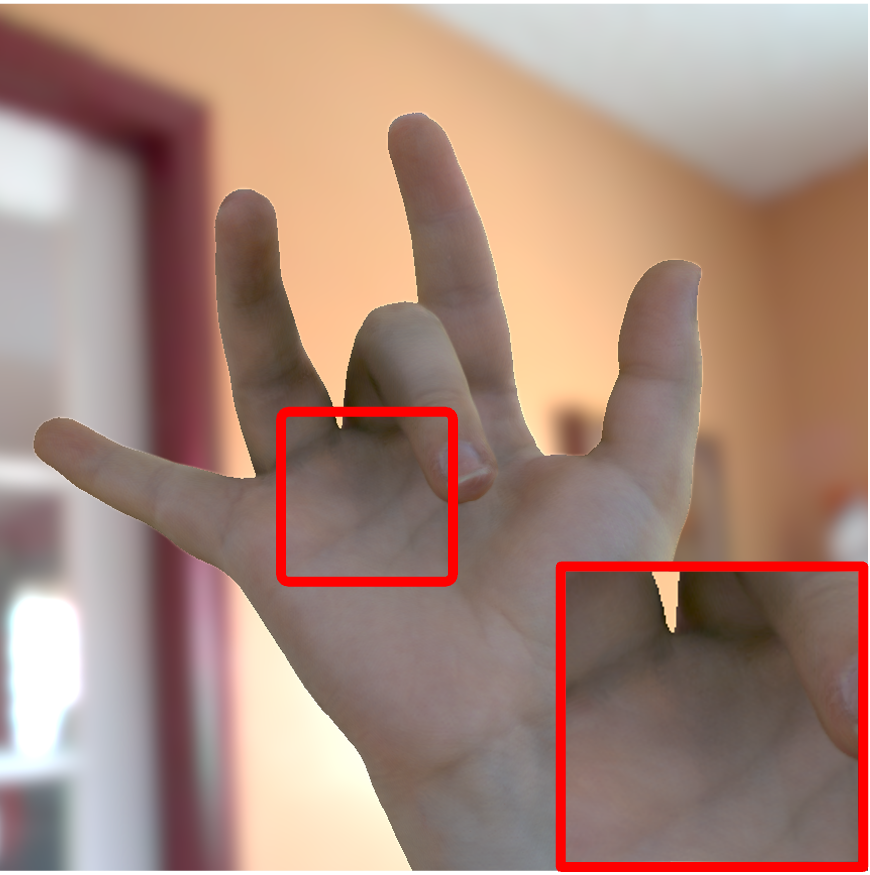}
       \caption*{w/o Visibility}
    \end{minipage}
    \begin{minipage}{0.19\textwidth}
       \centering
       \includegraphics[width=1.0\textwidth]{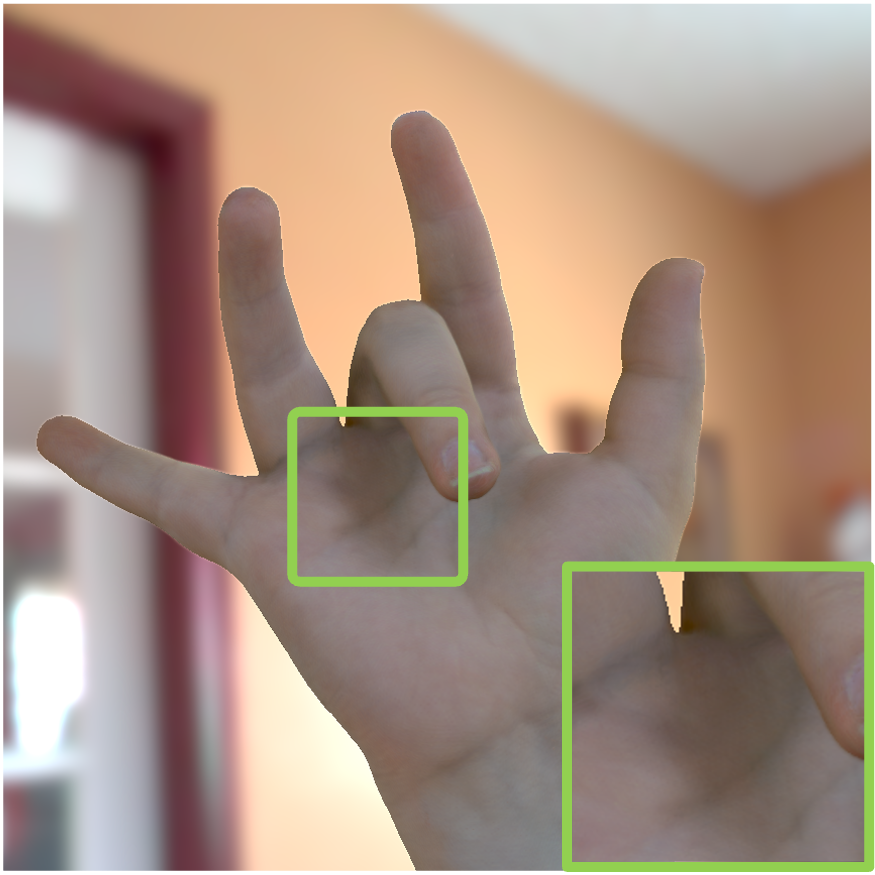}
       \caption*{w/o Specular Feature}
    \end{minipage}
    \begin{minipage}{0.19\textwidth}
       \centering
       \includegraphics[width=1.0\textwidth]{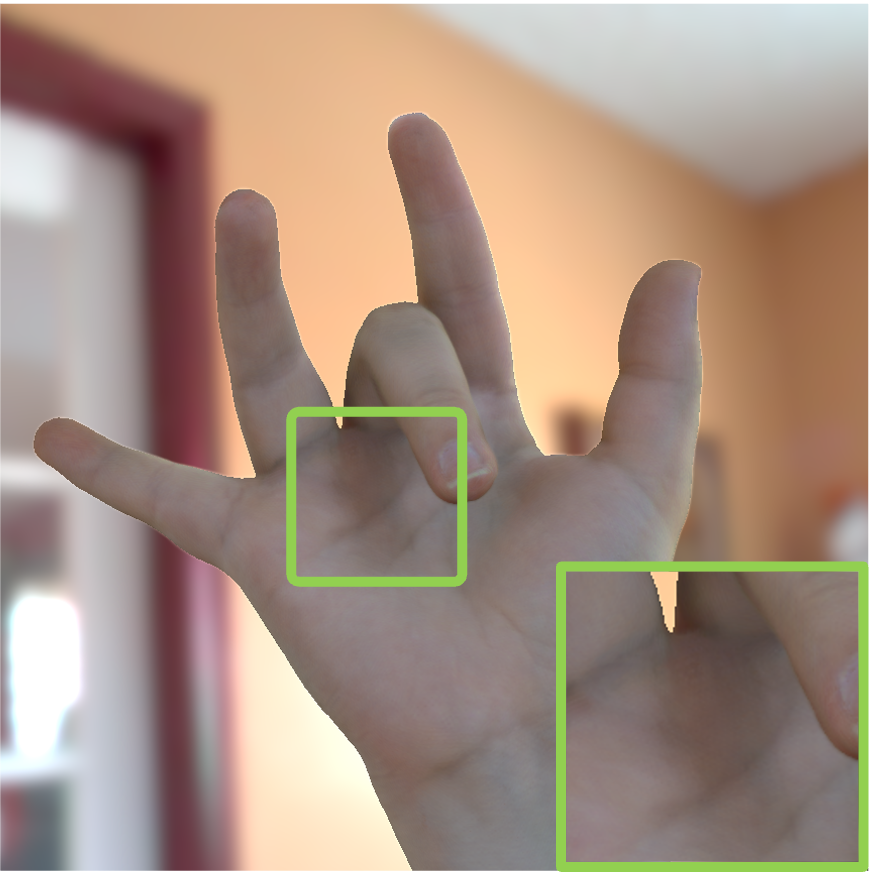}
       \caption*{Ours}
    \end{minipage}
    \begin{minipage}{0.19\textwidth}
       \centering
       \includegraphics[width=1.0\textwidth]{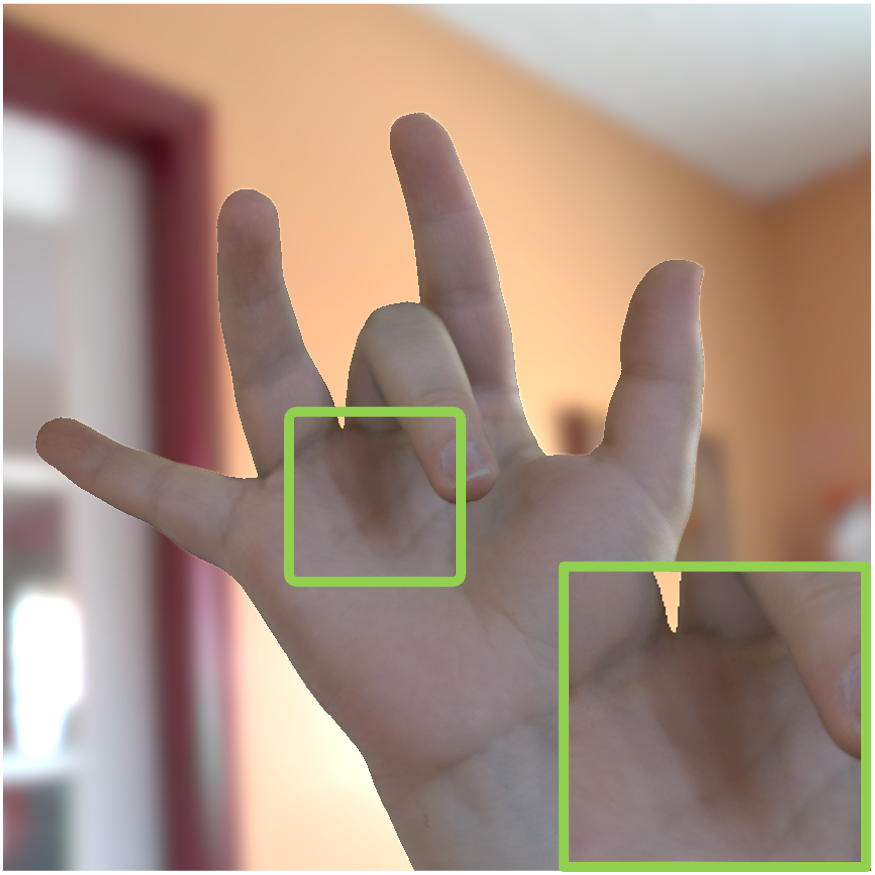}
       \caption*{Ground-Truth (Teacher)}
    \end{minipage}
    }
    \vspace{-0.3cm}
    \caption{\small{\textbf{Qualitative comparison of the student model.} A model-based relighting method~\cite{bi2021deep} fails to reproduce the precise color and fine-grained shading effects. Our model without visibility integration or specular features also lacks pose-dependent shadow or specular highlight respectively. In contrast, our full model successfully reproduces both effects. The green and red bounding boxes denote success and failure cases respectively.}}
    \label{fig:student_images_comparison}
    \vspace{-0.5cm}
\end{figure*}

\noindent
{\bf   Comparison to Bottleneck Light Conditioning.}
To evaluate the effectiveness of our spatially aligned lighting features, we compare against a bottleneck light representation with a hyper-network proposed in~\cite{bi2021deep}. 
For fair comparison, we replace our illumination encoding with their hyper-network while retaining everything else. 
\Cref{fig:student_images_comparison} shows that a bottleneck-based light encoding fails to match the overall intensity compared to the ground-truth. Moreover, it lacks fine-grained illumination effects such as reflection from grazing angles and soft shadows.
In contrast, our spatially aligned representation even without the proposed visibility integration significantly improves the fidelity of reconstruction. 
This observation is also strongly supported by our quantitative evaluation as shown in~\Cref{tab:student_evaluation}.

\noindent
{\bf   Ablation on Visibility Integration.}
We also evaluate the effectiveness of the proposed efficient visibility integration for computing features.
\Cref{fig:student_images_comparison} and \Cref{tab:student_evaluation} show that, compared to the model without the visibility integration, our full model achieves more faithful reconstruction of shadows even for novel poses and illuminations.

\noindent
{\bf   Ablation on Specular Features.}
We also validate the importance of specular features in the student model. 
\Cref{fig:student_images_comparison} and \Cref{tab:student_evaluation} illustrate that our specular feature provides sufficient information to reproduce specular highlight despite having the feature computed on a coarse proxy geometry.
This suggests that
spatially aligned lighting features are essential for achieving
generalizable neural relighting.

\noindent
{\bf   Runtime Analysis.}
One of our key contributions is the efficient rendering speed. 
While the teacher model takes approximately $30$ seconds to generate a texture with an envmap by aggregating over $512(=16\times32)$ light sources,
the student model achieves $48$ fps ($21$ ms) for a single hand and $31$ FPS ($32$ ms) for two hands on NVIDIA V100.

\vspace{-0.3cm}
\section{Discussion and Future Work}
We introduced the first model-based neural relighting for articulated hand models to enable photorealistic rendering of personalized hands under various illuminations in real-time. 
We successfully extend the teacher-student framework to build articulated models using a mesh-volumetric hybrid representation from multi-view light-stage capture data.
The hybrid representation allows us to use a coarse mesh to efficiently compute physics-inspired light features as input conditioning for the proposed student model.
Our experiments show that the spatially aligned light representation and explicit visibility integration are critical for highly generalizable relighting to novel poses and illuminations.

\vspace{-0.3cm}
\paragraph{Limitations and Future Work.}
Our student model currently does not support inter-reflection by other nearby objects due to far-field light assumption, which can be partially addressed by taking surroundings as a spatially varying envmap.
Future work also includes extending the proposed approach to clothed bodies, where computing visibility at a coarse mesh would not be sufficient for recovering fine-level shading caused by clothing deformations. 
Another exciting direction is to build a universal relightable hand model that spans inter-subject variations. As demonstrated by recent work on face modeling~\cite{cao2022authentic}, such a universal model would enable adaptation from in-the-wild inputs.

{\small
\bibliographystyle{ieee_fullname}
\bibliography{egbib}

\begin{thebibliography}{10}\itemsep=-1pt

\bibitem{alldieck2021imghum}
Thiemo Alldieck, Hongyi Xu, and Cristian Sminchisescu.
\newblock imghum: Implicit generative models of 3d human shape and articulated
  pose.
\newblock In {\em ICCV}, 2021.

\bibitem{bagautdinov2021driving}
Timur Bagautdinov, Chenglei Wu, Tomas Simon, Fabian Prada, Takaaki Shiratori,
  Shih-En Wei, Weipeng Xu, Yaser Sheikh, and Jason Saragih.
\newblock Driving-signal aware full-body avatars.
\newblock {\em TOG}.

\bibitem{ballan2012motion}
Luca Ballan, Aparna Taneja, J{\"u}rgen Gall, Luc~Van Gool, and Marc Pollefeys.
\newblock Motion capture of hands in action using discriminative salient
  points.
\newblock In {\em ECCV}, 2012.

\bibitem{bi2021deep}
Sai Bi, Stephen Lombardi, Shunsuke Saito, Tomas Simon, Shih-En Wei, Kevyn
  Mcphail, Ravi Ramamoorthi, Yaser Sheikh, and Jason Saragih.
\newblock Deep relightable appearance models for animatable faces.
\newblock {\em TOG}, 2021.

\bibitem{cao2022authentic}
Chen Cao, Tomas Simon, Jin~Kyu Kim, Gabe Schwartz, Michael Zollhoefer,
  Shun-Suke Saito, Stephen Lombardi, Shih-En Wei, Danielle Belko, Shoou-I Yu,
  et~al.
\newblock Authentic volumetric avatars from a phone scan.
\newblock {\em TOG}.

\bibitem{chen2022relighting}
Zhaoxi Chen and Ziwei Liu.
\newblock Relighting4d: Neural relightable human from videos.
\newblock In {\em ECCV}, 2022.

\bibitem{corona2022lisa}
Enric Corona, Tomas Hodan, Minh Vo, Francesc Moreno-Noguer, Chris Sweeney,
  Richard Newcombe, and Lingni Ma.
\newblock Lisa: Learning implicit shape and appearance of hands.
\newblock In {\em CVPR}, 2022.

\bibitem{de2011model}
Martin de La~Gorce, David~J Fleet, and Nikos Paragios.
\newblock Model-based 3d hand pose estimation from monocular video.
\newblock {\em PAMI}, 2011.

\bibitem{debevec2000acquiring}
Paul Debevec, Tim Hawkins, Chris Tchou, Haarm-Pieter Duiker, Westley Sarokin,
  and Mark Sagar.
\newblock Acquiring the reflectance field of a human face.
\newblock In {\em SIGGRAPH}, 2000.

\bibitem{deng2019neural}
Boyang Deng, JP Lewis, Timothy Jeruzalski, Gerard Pons-Moll, Geoffrey Hinton,
  Mohammad Norouzi, and Andrea Tagliasacchi.
\newblock Neural articulated shape approximation.
\newblock In {\em ECCV}, 2020.

\bibitem{Fyffe14}
Graham Fyffe, Andrew Jones, Oleg Alexander, Ryosuke Ichikari, and Paul Debevec.
\newblock Driving high-resolution facial scans with video performance capture.
\newblock {\em TOG}, 2014.

\bibitem{gall2009motion}
Juergen Gall, Carsten Stoll, Edilson De~Aguiar, Christian Theobalt, Bodo
  Rosenhahn, and Hans-Peter Seidel.
\newblock Motion capture using joint skeleton tracking and surface estimation.
\newblock In {\em CVPR}, 2009.

\bibitem{gardner2017learning}
Marc-Andr{\'e} Gardner, Kalyan Sunkavalli, Ersin Yumer, Xiaohui Shen, Emiliano
  Gambaretto, Christian Gagn{\'e}, and Jean-Fran{\c{c}}ois Lalonde.
\newblock Learning to predict indoor illumination from a single image.
\newblock {\em TOG}, 2017.

\bibitem{Ghosh11}
Abhijeet Ghosh, Graham Fyffe, Borom Tunwattanapong, Jay Busch, Xueming Yu, and
  Paul Debevec.
\newblock Multiview face capture using polarized spherical gradient
  illumination.
\newblock {\em TOG}, 2011.

\bibitem{guo2019relightables}
Kaiwen Guo, Peter Lincoln, Philip Davidson, Jay Busch, Xueming Yu, Matt Whalen,
  Geoff Harvey, Sergio Orts-Escolano, Rohit Pandey, Jason Dourgarian, et~al.
\newblock The relightables: Volumetric performance capture of humans with
  realistic relighting.
\newblock {\em TOG}, 2019.

\bibitem{hasson2019learning}
Yana Hasson, Gul Varol, Dimitrios Tzionas, Igor Kalevatykh, Michael~J Black,
  Ivan Laptev, and Cordelia Schmid.
\newblock Learning joint reconstruction of hands and manipulated objects.
\newblock In {\em CVPR}, 2019.

\bibitem{face-relighting-with-geometrically-consistent-shadows}
Andrew Hou, Michel Sarkis, Ning Bi, Yiying Tong, and Xiaoming Liu.
\newblock Face relighting with geometrically consistent shadows.
\newblock In {\em CVPR}, 2022.

\bibitem{towards-high-fidelity-face-relighting-with-realistic-shadows}
Andrew Hou, Ze Zhang, Michel Sarkis, Ning Bi, Yiying Tong, and Xiaoming Liu.
\newblock Towards high fidelity face relighting with realistic shadows.
\newblock In {\em CVPR}, 2021.

\bibitem{Ji_undated-vp}
Chaonan Ji, Tao Yu, Kaiwen Guo, Jingxin Liu, and Yebin Liu.
\newblock Geometry-aware single-image full-body human relighting.
\newblock In {\em ECCV}, 2022.

\bibitem{kanamori2018relight}
Yoshihiro Kanamori and Yuki Endo.
\newblock Relighting humans: Occlusion-aware inverse rendering for full-body
  human images.
\newblock {\em SIGGRAPH Asia}, 2018.

\bibitem{karunratanakul2021halo}
Korrawe Karunratanakul, Adrian Spurr, Zicong Fan, Otmar Hilliges, and Siyu
  Tang.
\newblock A skeleton-driven neural occupancy representation for articulated
  hands.
\newblock In {\em 3DV}, 2021.

\bibitem{kingma:adam}
Diederick~P Kingma and Jimmy Ba.
\newblock Adam: A method for stochastic optimization.
\newblock In {\em ICLR}, 2015.

\bibitem{Lagunas:2021}
Manuel Lagunas, Xin Sun, Jimei Yang, Ruben Villegas, Jianming Zhang, Zhixin
  Shu, Belen Masia, and Diego Gutiérrez.
\newblock Single-image full-body human relighting.
\newblock In {\em Eurographics Symposium on Rendering}, 2021.

\bibitem{li2022eyenerf}
Gengyan Li, Abhimitra Meka, Franziska Mueller, Marcel~C. Buehler, Otmar
  Hilliges, and Thabo Beeler.
\newblock Eyenerf: A hybrid representation for photorealistic synthesis,
  animation and relighting of human eyes.
\newblock {\em SIGGRAPH}, 2022.

\bibitem{li2022interacting}
Mengcheng Li, Liang An, Hongwen Zhang, Lianpeng Wu, Feng Chen, Tao Yu, and
  Yebin Liu.
\newblock Interacting attention graph for single image two-hand reconstruction.
\newblock In {\em CVPR}, 2022.

\bibitem{li2019rethinking}
Wenbo Li, Zhicheng Wang, Binyi Yin, Qixiang Peng, Yuming Du, Tianzi Xiao, Gang
  Yu, Hongtao Lu, Yichen Wei, and Jian Sun.
\newblock Rethinking on multi-stage networks for human pose estimation.
\newblock {\em arXiv:1901.00148}, 2019.

\bibitem{li2022nimble}
Yuwei Li, Longwen Zhang, Zesong Qiu, Yingwenqi Jiang, Nianyi Li, Yuexin Ma,
  Yuyao Zhang, Lan Xu, and Jingyi Yu.
\newblock Nimble: A non-rigid hand model with bones and muscles.
\newblock {\em SIGGRAPH}, 2022.

\bibitem{lokovic2000deep}
Tom Lokovic and Eric Veach.
\newblock Deep shadow maps.
\newblock In {\em SIGGRAPH}, 2000.

\bibitem{lombardi2018deep}
Stephen Lombardi, Jason Saragih, Tomas Simon, and Yaser Sheikh.
\newblock Deep appearance models for face rendering.
\newblock {\em TOG}, 2018.

\bibitem{lombardi2021mvp}
Stephen Lombardi, Tomas Simon, Gabriel Schwartz, Michael Zollhoefer, Yaser
  Sheikh, and Jason Saragih.
\newblock Mixture of volumetric primitives for efficient neural rendering.
\newblock {\em TOG}, 2021.

\bibitem{Ma07}
Wan-Chun Ma, Tim Hawkins, Pieter Peers, Charles-Felix Chabert, Malte Weiss, and
  Paul Debevec.
\newblock Rapid acquisition of specular and diffuse normal maps from polarized
  spherical gradient illumination.
\newblock In {\em EGSR}, 2007.

\bibitem{meka2019deep}
Abhimitra Meka, Christian Haene, Rohit Pandey, Michael Zollh{\"o}fer, Sean
  Fanello, Graham Fyffe, Adarsh Kowdle, Xueming Yu, Jay Busch, Jason
  Dourgarian, et~al.
\newblock Deep reflectance fields: high-quality facial reflectance field
  inference from color gradient illumination.
\newblock {\em TOG}.

\bibitem{moon2018v2v}
Gyeongsik Moon, Ju~Yong Chang, and Kyoung~Mu Lee.
\newblock V2v-posenet: Voxel-to-voxel prediction network for accurate 3d hand
  and human pose estimation from a single depth map.
\newblock In {\em CVPR}, 2018.

\bibitem{moon2020i2l}
Gyeongsik Moon and Kyoung~Mu Lee.
\newblock I2l-meshnet: Image-to-lixel prediction network for accurate 3d human
  pose and mesh estimation from a single rgb image.
\newblock In {\em ECCV}, 2020.

\bibitem{moon2020deephandmesh}
Gyeongsik Moon, Takaaki Shiratori, and Kyoung~Mu Lee.
\newblock Deephandmesh: A weakly-supervised deep encoder-decoder framework for
  high-fidelity hand mesh modeling.
\newblock In {\em ECCV}, 2020.

\bibitem{mueller2018ganerated}
Franziska Mueller, Florian Bernard, Oleksandr Sotnychenko, Dushyant Mehta,
  Srinath Sridhar, Dan Casas, and Christian Theobalt.
\newblock Ganerated hands for real-time 3d hand tracking from monocular rgb.
\newblock In {\em CVPR}, 2018.

\bibitem{mueller2019real}
Franziska Mueller, Micah Davis, Florian Bernard, Oleksandr Sotnychenko, Mickeal
  Verschoor, Miguel~A Otaduy, Dan Casas, and Christian Theobalt.
\newblock Real-time pose and shape reconstruction of two interacting hands with
  a single depth camera.
\newblock {\em TOG}, 2019.

\bibitem{mueller2017real}
Franziska Mueller, Dushyant Mehta, Oleksandr Sotnychenko, Srinath Sridhar, Dan
  Casas, and Christian Theobalt.
\newblock Real-time hand tracking under occlusion from an egocentric rgb-d
  sensor.
\newblock In {\em ICCV}, 2017.

\bibitem{DBLP:conf/cvpr/NestmeyerLML20}
Thomas Nestmeyer, Jean{-}Fran{\c{c}}ois Lalonde, Iain~A. Matthews, and
  Andreas~M. Lehrmann.
\newblock Learning physics-guided face relighting under directional light.
\newblock In {\em CVPR}, 2020.

\bibitem{2021narf}
Atsuhiro Noguchi, Xiao Sun, Stephen Lin, and Tatsuya Harada.
\newblock Neural articulated radiance field.
\newblock In {\em ICCV}, 2021.

\bibitem{oikonomidis2011efficient}
Iason Oikonomidis, Nikolaos Kyriazis, and Antonis~A Argyros.
\newblock Efficient model-based 3d tracking of hand articulations using kinect.
\newblock In {\em BMVC}, 2011.

\bibitem{Pandey:2021}
Rohit Pandey, Sergio Orts-Escolano, Chloe LeGendre, Christian Haene, Sofien
  Bouaziz, Christoph Rhemann, Paul Debevec, and Sean Fanello.
\newblock Total relighting: Learning to relight portraits for background
  replacement.
\newblock In {\em TOG}, 2021.

\bibitem{parker2010optix}
Steven~G Parker, James Bigler, Andreas Dietrich, Heiko Friedrich, Jared
  Hoberock, David Luebke, David McAllister, Morgan McGuire, Keith Morley,
  Austin Robison, et~al.
\newblock Optix: a general purpose ray tracing engine.
\newblock {\em TOG}.

\bibitem{peng2021neural}
Sida Peng, Yuanqing Zhang, Yinghao Xu, Qianqian Wang, Qing Shuai, Hujun Bao,
  and Xiaowei Zhou.
\newblock Neural body: Implicit neural representations with structured latent
  codes for novel view synthesis of dynamic humans.
\newblock In {\em CVPR}, 2021.

\bibitem{HTML_eccv2020}
Neng Qian, Jiayi Wang, Franziska Mueller, Florian Bernard, Vladislav Golyanik,
  and Christian Theobalt.
\newblock {HTML: A Parametric Hand Texture Model for 3D Hand Reconstruction and
  Personalization}.
\newblock In {\em ECCV}, 2020.

\bibitem{rehg1994visual}
James~M Rehg and Takeo Kanade.
\newblock Visual tracking of high dof articulated structures: an application to
  human hand tracking.
\newblock In {\em ECCV}, 1994.

\bibitem{remelli2022dva}
Edoardo Remelli, Timur Bagautdinov, Shunsuke Saito, Chenglei Wu, Tomas Simon,
  Shih-En Wei, Kaiwen Guo, Zhe Cao, Fabian Prada, Jason Saragih, and Yaser
  Sheikh.
\newblock Drivable volumetric avatars using texel-aligned features.
\newblock 2022.

\bibitem{MANO:SIGGRAPHASIA:2017}
Javier Romero, Dimitrios Tzionas, and Michael~J. Black.
\newblock Embodied hands: Modeling and capturing hands and bodies together.
\newblock {\em SIGGRAPH Asia}, 2017.

\bibitem{10.1007/978-3-319-24574-4_28}
Olaf Ronneberger, Philipp Fischer, and Thomas Brox.
\newblock U-net: Convolutional networks for biomedical image segmentation.
\newblock In Nassir Navab, Joachim Hornegger, William~M. Wells, and
  Alejandro~F. Frangi, editors, {\em MICCAI}, 2015.

\bibitem{saito2021scanimate}
Shunsuke Saito, Jinlong Yang, Qianli Ma, and Michael~J Black.
\newblock Scanimate: Weakly supervised learning of skinned clothed avatar
  networks.
\newblock In {\em CVPR}, 2021.

\bibitem{sfsnetSengupta18}
Soumyadip Sengupta, Angjoo Kanazawa, Carlos~D. Castillo, and David~W. Jacobs.
\newblock Sfsnet: Learning shape, refectance and illuminance of faces in the
  wild.
\newblock In {\em CVPR}, 2018.

\bibitem{simon2017hand}
Tomas Simon, Hanbyul Joo, Iain Matthews, and Yaser Sheikh.
\newblock Hand keypoint detection in single images using multiview
  bootstrapping.
\newblock In {\em CVPR}, 2017.

\bibitem{smith2020constraining}
Breannan Smith, Chenglei Wu, He Wen, Patrick Peluse, Yaser Sheikh, Jessica~K
  Hodgins, and Takaaki Shiratori.
\newblock Constraining dense hand surface tracking with elasticity.
\newblock {\em TOG}.

\bibitem{sridhar2015fast}
Srinath Sridhar, Franziska Mueller, Antti Oulasvirta, and Christian Theobalt.
\newblock Fast and robust hand tracking using detection-guided optimization.
\newblock In {\em CVPR}, 2015.

\bibitem{sridhar2013interactive}
Srinath Sridhar, Antti Oulasvirta, and Christian Theobalt.
\newblock Interactive markerless articulated hand motion tracking using rgb and
  depth data.
\newblock In {\em ICCV}, 2013.

\bibitem{sun2019single}
Tiancheng Sun, Jonathan~T. Barron, Yun-Ta Tsai, Zexiang Xu, Xueming Yu, Graham
  Fyffe, Christoph Rhemann, Jay Busch, Paul Debevec, and Ravi Ramamoorthi.
\newblock Single image portrait relighting.
\newblock {\em TOG}, 2019.

\bibitem{sun2020light}
Tiancheng Sun, Zexiang Xu, Xiuming Zhang, Sean Fanello, Christoph Rhemann, Paul
  Debevec, Yun-Ta Tsai, Jonathan~T Barron, and Ravi Ramamoorthi.
\newblock Light stage super-resolution: continuous high-frequency relighting.
\newblock {\em TOG}, 2020.

\bibitem{tagliasacchi2015robust}
Andrea Tagliasacchi, Matthias Schr{\"o}der, Anastasia Tkach, Sofien Bouaziz,
  Mario Botsch, and Mark Pauly.
\newblock Robust articulated-icp for real-time hand tracking.
\newblock In {\em Computer graphics forum}, 2015.

\bibitem{tkach2016sphere}
Anastasia Tkach, Mark Pauly, and Andrea Tagliasacchi.
\newblock Sphere-meshes for real-time hand modeling and tracking.
\newblock {\em TOG}, 2016.

\bibitem{tzionas2016capturing}
Dimitrios Tzionas, Luca Ballan, Abhilash Srikantha, Pablo Aponte, Marc
  Pollefeys, and Juergen Gall.
\newblock Capturing hands in action using discriminative salient points and
  physics simulation.
\newblock {\em IJCV}, 2016.

\bibitem{10.5555/2383847.2383874}
Bruce Walter, Stephen~R. Marschner, Hongsong Li, and Kenneth~E. Torrance.
\newblock Microfacet models for refraction through rough surfaces.
\newblock In {\em ESGR}, 2007.

\bibitem{wang2019hand}
Bohan Wang, George Matcuk, and Jernej Barbi{\v{c}}.
\newblock Hand modeling and simulation using stabilized magnetic resonance
  imaging.
\newblock {\em TOG}.

\bibitem{wang2020single}
Zhibo Wang, Xin Yu, Ming Lu, Quan Wang, Chen Qian, and Feng Xu.
\newblock Single image portrait relighting via explicit multiple reflectance
  channel modeling.
\newblock {\em TOG}, 2020.

\bibitem{wenger2005performance}
Andreas Wenger, Andrew Gardner, Chris Tchou, Jonas Unger, Tim Hawkins, and Paul
  Debevec.
\newblock Performance relighting and reflectance transformation with
  time-multiplexed illumination.
\newblock {\em TOG}, 2005.

\bibitem{weyrich2006analysis}
Tim Weyrich, Wojciech Matusik, Hanspeter Pfister, Bernd Bickel, Craig Donner,
  Chien Tu, Janet McAndless, Jinho Lee, Addy Ngan, Henrik~Wann Jensen, et~al.
\newblock Analysis of human faces using a measurement-based skin reflectance
  model.
\newblock {\em TOG}, 2006.

\bibitem{xu2018deep}
Zexiang Xu, Kalyan Sunkavalli, Sunil Hadap, and Ravi Ramamoorthi.
\newblock Deep image-based relighting from optimal sparse samples.
\newblock {\em TOG}, 2018.

\bibitem{yamaguchi2018high}
Shugo Yamaguchi, Shunsuke Saito, Koki Nagano, Yajie Zhao, Weikai Chen, Kyle
  Olszewski, Shigeo Morishima, and Hao Li.
\newblock High-fidelity facial reflectance and geometry inference from an
  unconstrained image.
\newblock {\em TOG}, 2018.

\bibitem{yeh2022learning}
Yu-Ying Yeh, Koki Nagano, Sameh Khamis, Jan Kautz, Ming-Yu Liu, and Ting-Chun
  Wang.
\newblock Learning to relight portrait images via a virtual light stage and
  synthetic-to-real adaptation.
\newblock {\em SIGGRAPH Asia}, 2022.

\bibitem{zhang2021nlt}
Xiuming Zhang, Sean Fanello, Yun-Ta Tsai, Tiancheng Sun, Tianfan Xue, Rohit
  Pandey, Sergio Orts-Escolano, Philip Davidson, Christoph Rhemann, Paul
  Debevec, Jonathan~T. Barron, Ravi Ramamoorthi, and William~T. Freeman.
\newblock Neural light transport for relighting and view synthesis.
\newblock {\em SIGGRAPH}, 2021.

\bibitem{Zheng:2022:SOH}
Mianlun Zheng, Bohan Wang, Jingtao Huang, and Jernej Barbi\v{c}.
\newblock Simulation of hand anatomy using medical imaging.
\newblock {\em SIGGRAPH Asia}, 2022.

\bibitem{zhou2020monocular}
Yuxiao Zhou, Marc Habermann, Weipeng Xu, Ikhsanul Habibie, Christian Theobalt,
  and Feng Xu.
\newblock Monocular real-time hand shape and motion capture using multi-modal
  data.
\newblock In {\em CVPR}, 2020.

\end{thebibliography}
}

\clearpage
\appendix

\begin{table*}[t]
    \centering
    \setlength\extrarowheight{1pt}
    \small{
        \scalebox{0.85}{
        \begin{tabular}{c||ccc|ccc|ccc|ccc}
            & \multicolumn{6}{c|}{Subject 1} & \multicolumn{6}{c}{Subject 2} \\ \hline
            & \multicolumn{3}{c|}{MSE ($\times 10^{-3}$) $\downarrow$} & \multicolumn{3}{c|}{SSIM $\uparrow$} & \multicolumn{3}{c|}{MSE ($\times 10^{-3}$) $\downarrow$} & \multicolumn{3}{c}{SSIM $\uparrow$} \\ \hline
            & Right & Left & Both & Right & Left & Both & Right & Left & Both & Right & Left & Both \\ \hline 
            GGX~\cite{10.5555/2383847.2383874,chen2022relighting} & 15.2151 & 19.5409 & 59.7985 & 0.9591 & 0.9579 & 0.9201 & 22.3159 & 30.0019 & 65.5644 & 0.9145 & 0.9231 & 0.8695 \\ \hline
            Ours                                   & \textbf{4.9126} & \textbf{5.8608} & \textbf{15.7589} & \textbf{0.9790} & \textbf{0.9805} & \textbf{0.9536} & \textbf{8.8205} & \textbf{7.9357} & \textbf{22.3559} & \textbf{0.9541} & \textbf{0.9559} & \textbf{0.9075} \\
        \end{tabular} 
        }
    }
    \vspace{-0.2cm} 
    \caption{\textbf{Quantitative comparison of the teacher model.} We measure the MSE and SSIM metrics on the right, left, and two-hand sequences. The result shows that our method significantly outperforms the physically-based rendering baseline based on Relighting4D~\cite{chen2022relighting}.}
    \label{tab:teacher_evaluation}
\end{table*}

\section{Network Architecture}
In this section, we provide the network architecture details and hyperparameters for the teacher and student models. In addition, we explain the modifications we made to DRAM~\cite{bi2021deep} for a fair comparison to our proposed method.

\subsection{Teacher Model}
The joint feature encoder $\mathcal{J}_t \left(\theta\right)$ of the teacher model first repeats and tiles the hand pose $\theta \in \mathbb{R}^{25}$ across the UV space and $\theta^{\prime} \in \mathbb{R}^{25\times64\times64}$ is obtained
Given the hand pose feature $\theta^{\prime}$, a $2$-layer convolutional neural network (CNN) with channel sizes $(16, 64)$ outputs a joint feature $\mathcal{J}_t \left(\theta \right) \in \mathbb{R}^{64 \times 64 \times 64}$. $\mathcal{A}_{\text{OLAT}}$ adopts a U-Net~\cite{10.1007/978-3-319-24574-4_28} architecture
which takes $wS \times wS \times 7S$ per-light feature as an input. The U-Net encoder is a $4$-layer CNN with channel sizes $(7S, 64, 64, 64)$. Here, the joint feature is concatenated with the output feature from the U-Net encoder and passed to the U-Net decoder. The U-Net decoder is a $4$-layer CNN with a skip connection from the U-Net encoder, with channel sizes $(128, 128, 128, 4S)$. We use bilinear interpolation to downsample and upsample the features at each layer. The first $3S$ output channels encode raw RGB volumetric texture $\mathbf{T}_k^i$, and the last $S$ channels the shadow map $\mathbf{S}_k^i$. The OLAT texture $\mathbf{C}_k^i$ for the $i$-th light is computed as follows;

\begin{equation}
    \mathbf{C}_k^{i} = \sigma(\mathbf{S}_k^i) \left(\text{ReLU}\left(\lambda_s \mathbf{T}_k^{i}\right) + \lambda_b \right) \in \mathbb{R}^{S \times S \times 3S},
\end{equation}
where $\sigma(\cdot)$ is a sigmoid function, $\text{ReLU}(x) = \max(0, x)$, $\lambda_s$ is a scale parameter, and $\lambda_b$ is a bias parameter. In our experiments, we set $\lambda_s$ to $25$ and $\lambda_b$ to $100$. 

\subsection{Student Model}
The joint feature encoder $\mathcal{J}_s \left(\theta\right)$ of the student model is the same architecture as $\mathcal{J}_t \left(\theta\right)$. However, the student model only has a texture decoder network $\mathcal{A}_{\text{env}}$ which takes a $76(=3+9+64)\times128\times128$ feature as an input. The texture decoder is an $8$-layer CNN with channel sizes $(76, 256, 256, 128, 128, 64, 64, 32, 3S)$, outputting raw RGB volumetric texture. Given the output raw RGB texture $\mathbf{T}_k$, the relit texture $\mathbf{C}_k$ is expressed as

\begin{equation}
    \mathbf{C}_k = \text{ReLU}\left(\lambda_s \mathbf{T}_k\right) + \lambda_b \in \mathbb{R}^{S \times S \times 3S},
\end{equation}
where the same hyperparameters $\lambda_s$ and $\lambda_b$ are the same as for the teacher model.

\subsection{DRAM~\cite{bi2021deep}}
Unlike DRAM~\cite{bi2021deep}, our hand model is animated by the hand pose parameter instead of facial expressions. Therefore, we replace the latent code $z$ in their hyper-network architecture with the hand pose parameter $\theta$. In addition, we rotate the environment map to align with the coordinate system of the hand's root joint before it is given to the hyper-network.

\section{Comparison to GGX Microfacet BRDF~\cite{10.5555/2383847.2383874}}
Our teacher model directly estimates an OLAT rendering from localized light and view directions and shadow maps. An alternative approach is to adopt a physically-based BRDF model and render with the rendering equation. To validate the effectiveness of our neural rendering approach, we compare the teacher model against a physically-based rendering baseline that is heavily inspired by the recent success of Relighting4D~\cite{chen2022relighting}. This physically-based baseline predicts material properties such as albedo, roughness, and a localized normal map per voxel, and feeds these into GGX microfacet BRDF~\cite{10.5555/2383847.2383874} to obtain the RGB textures.

We adopt the same network architecture as the original teacher model with the joint feature encoder for deep shadow map and albedo estimation except for the first and output channel sizes. Concretely, we give the local light directions $\mathbf{F}^k_v$ and visibility map $\mathbf{V}(l_i)_k$ as an input to the shadow map U-Net to obtain the per-light shadow map $\mathbf{S}^i_k \in \mathbb{R}^{S \times S \times S}$. Also, we feed the average texture $\bar{\mathbf{T}}$ into the albedo U-Net to acquire the albedo $\mathbf{A}_k \in \mathbb{R}^{S \times S \times 3S}$. Since MVP~\cite{lombardi2021mvp} does not provide normal information, we learn normal and roughness decoders with the channel sizes $(128, 128, 128, 4S)$ and the same architecture as the teacher model. GGX microfacet BRDF~\cite{10.5555/2383847.2383874} takes the estimated albedo $\mathbf{A}_k$, roughness $\mathbf{R}_k \in \mathbb{R}^{S \times S \times S}$, and normal $\mathbf{N}_k \in \mathbb{R}^{S \times S \times 3S}$ as input to compute RGB texture $\mathbf{T}_k$. Using the shadow map and the raw RGB texture, the predicted texture of the $k$-th primitive is expressed simply by
\vspace{-0.3cm}
\begin{equation}
    \mathbf{C}_k^i = \sigma(\mathbf{S}_k^i) f \left(\mathbf{A}_k, \mathbf{R}_k, \mathbf{N}_k \right) = \sigma(\mathbf{S}_k^i) \mathbf{T}_k \in \mathbb{R}^{S \times S \times 3S},
\end{equation}
where $f\left(\mathbf{A}_k, \mathbf{R}_k, \mathbf{N}_k \right)$ is the element-wise GGX microfacet BRDF function. To train the network, the same hyperparameters and loss fuction are used.

\begin{figure}
    \centering
    \setlength\extrarowheight{1pt}
    \begin{minipage}{0.15\textwidth}
       \centering
       \includegraphics[width=1.0\textwidth]{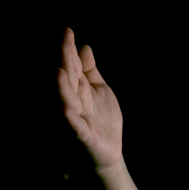}
       \caption*{Ground-Truth}
    \end{minipage}
    \begin{minipage}{0.15\textwidth}
       \centering
       \includegraphics[width=1.0\textwidth]{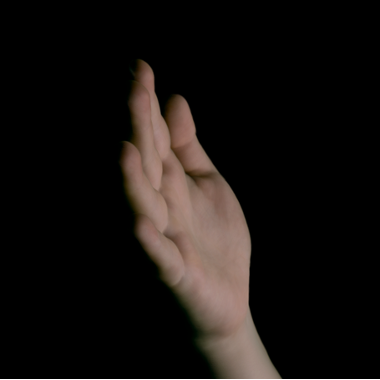}
       \caption*{Ours}
    \end{minipage}
    \begin{minipage}{0.15\textwidth}
       \centering
       \includegraphics[width=1.0\textwidth]{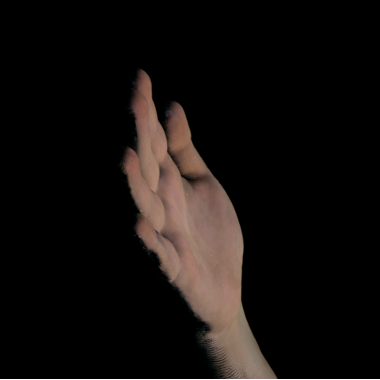}
       \caption*{GGX~\cite{chen2022relighting,10.5555/2383847.2383874}}
    \end{minipage}
    \caption{\textbf{Comparison with GGX microfacet BRDF model.} While the quality of relighting with physically-based relighting is bounded by the underlying BRDF model, our neural relighting model learns to synthesize photorealistic light transport effects without specifying material parameters.}
    \label{fig:teacher_relighting4d_subject1}
\end{figure}

\Cref{tab:teacher_evaluation} shows that our neural relighting outperforms the physically-based baseline by a large margin. This experimental result demonstrates that the quality of physically-based rendering is bounded by the underlying parameterization, and the GGX microfacet BRDF is unable to represent complex transmissive effects such as subsurface scattering. In contrast, the proposed neural relighting approach successfully models global light transport effects.
We show the qualitative comparison between our method and the physcally-based baseline in \Cref{fig:teacher_relighting4d_subject1}.

\begin{figure*}[htb!]
    \centering
    \includegraphics[width=\linewidth]{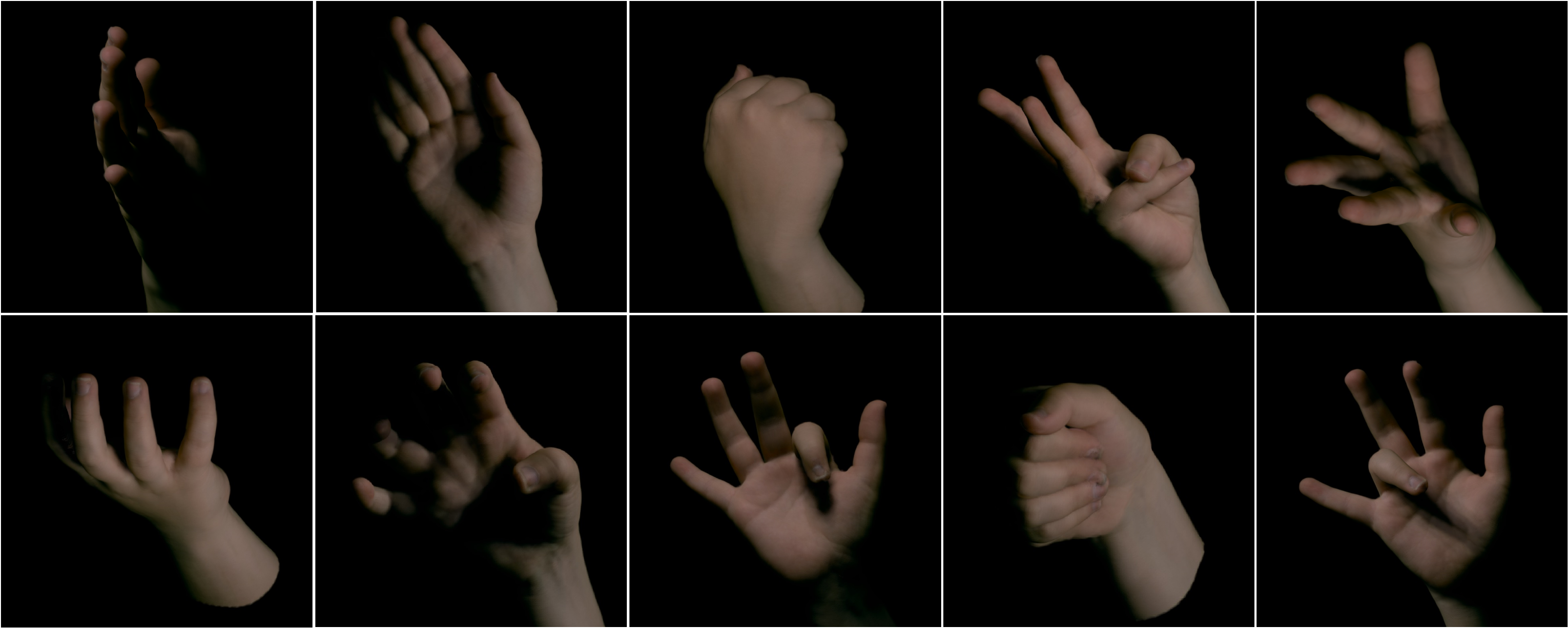}
    \caption{\textbf{Qualitative results of the teacher model with directional lighting.} Each image is generated by a randomly sampled pose and lighting condition from the test dataset of Subject 1.}
    \label{fig:teacher_directional_subject1}
\end{figure*}

\begin{figure*}[htb!]
    \centering
    \includegraphics[width=\linewidth]{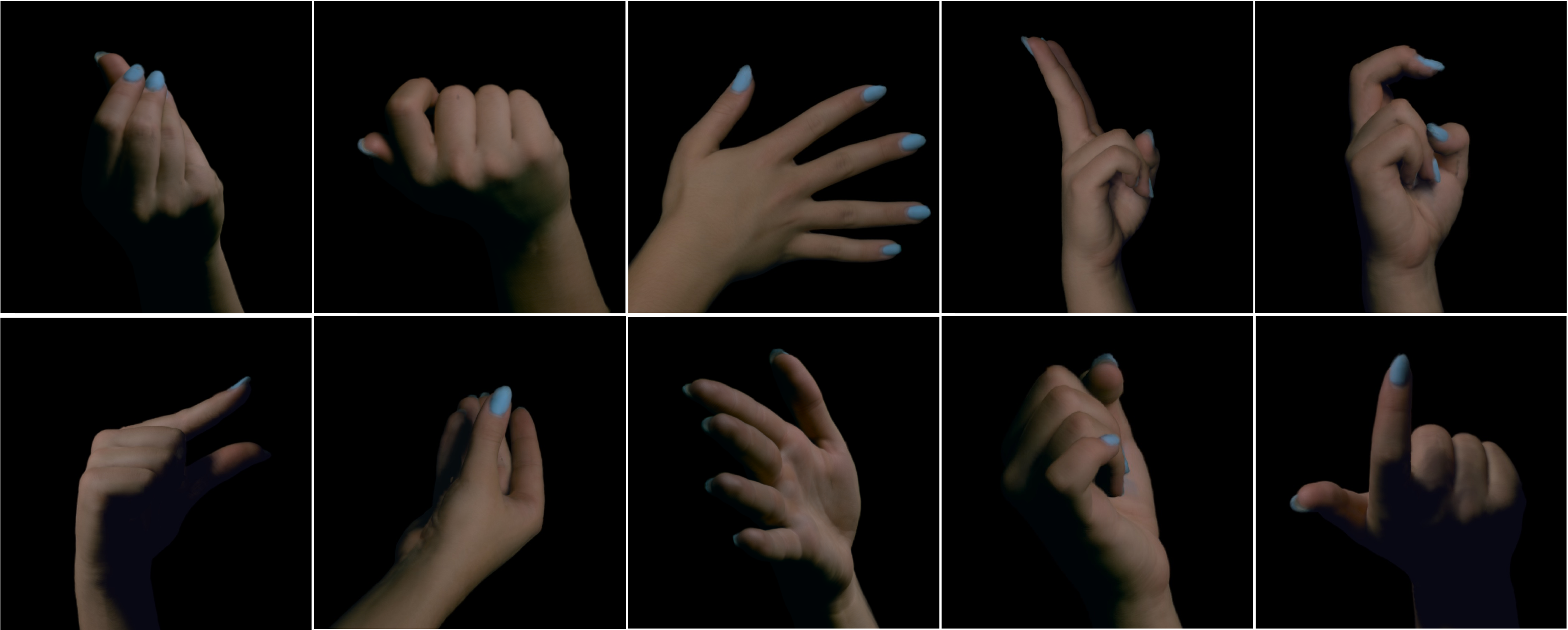}
    \caption{\textbf{Qualitative results of the teacher model with directional lighting.} Each image is generated by a randomly sampled pose and lighting condition from the test dataset of Subject 2.}
    \label{fig:teacher_directional_subject2}
\end{figure*}

\begin{figure*}[htb!]
    \centering
    \includegraphics[width=\linewidth]{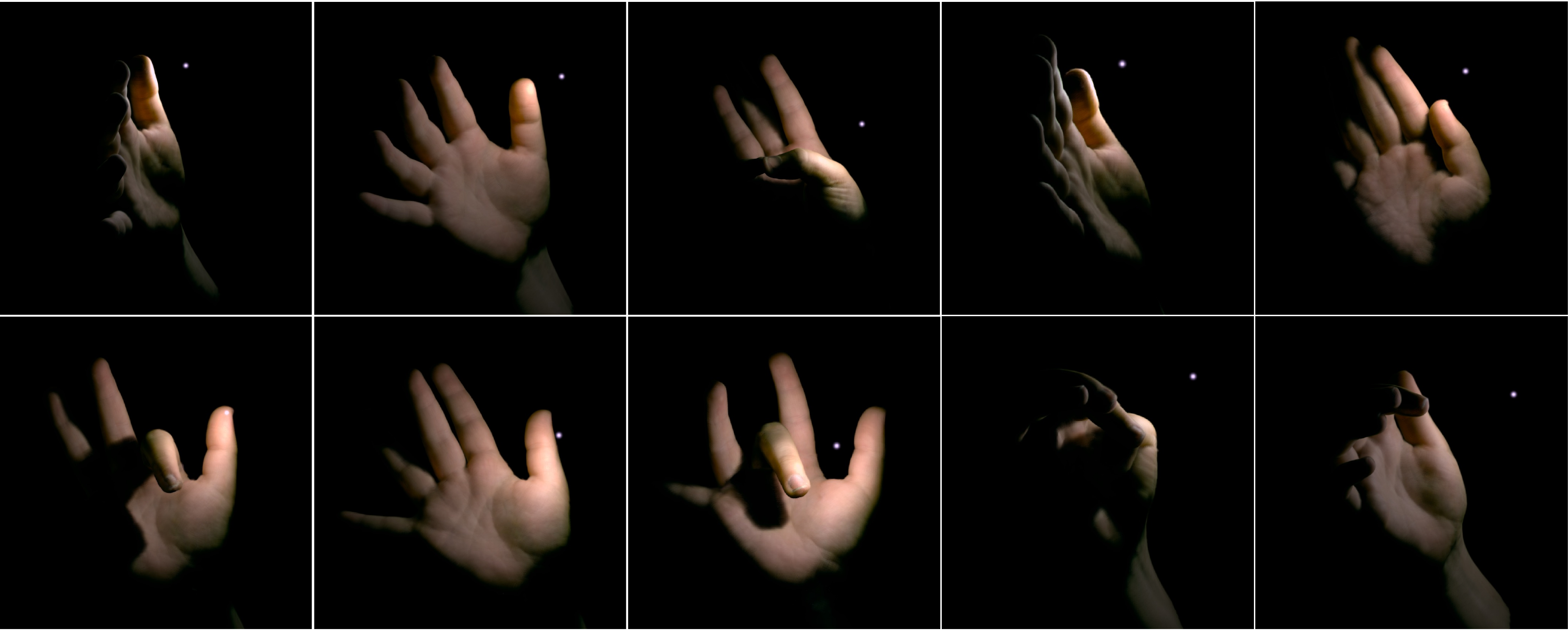}
    \caption{\textbf{Qualitative results of the teacher model with near-field lighting.} We sample the poses from the validation dataset of Subject 1. Note that we set the light locations around the fingers manually and the white dot represents the the projected light location in the 2D image space. }
    \label{fig:teacher_nearfield_subject1}
\end{figure*}

\begin{figure*}[htb!]
    \centering
    \includegraphics[width=\linewidth]{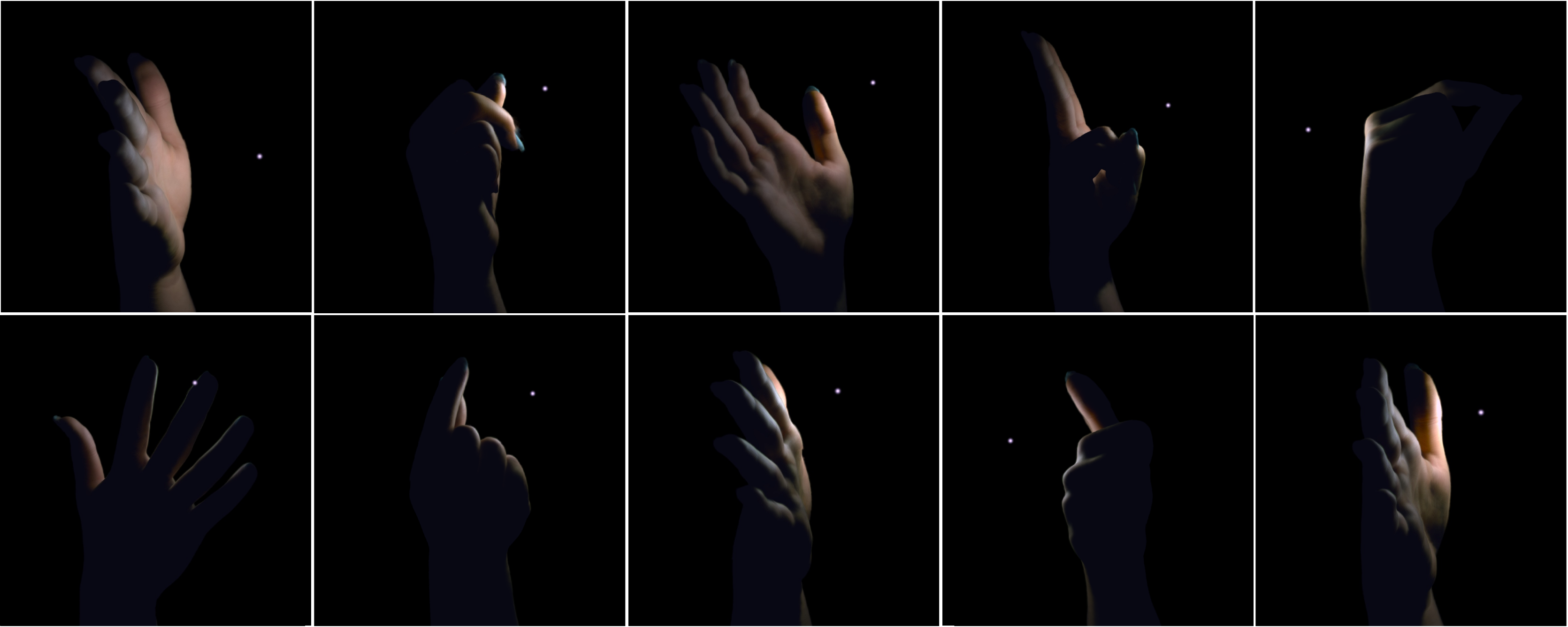}
    \caption{\textbf{Qualitative results of the teacher model with near-field lighting.} We sample the poses from the validation dataset of Subject 2. Note that we set the light locations around the fingers manually and the white dot represents the the projected light location in the 2D image space. }
    \label{fig:teacher_nearfield_subject2}
\end{figure*}

\begin{figure*}[htb!]
    \centering
    \includegraphics[width=\linewidth]{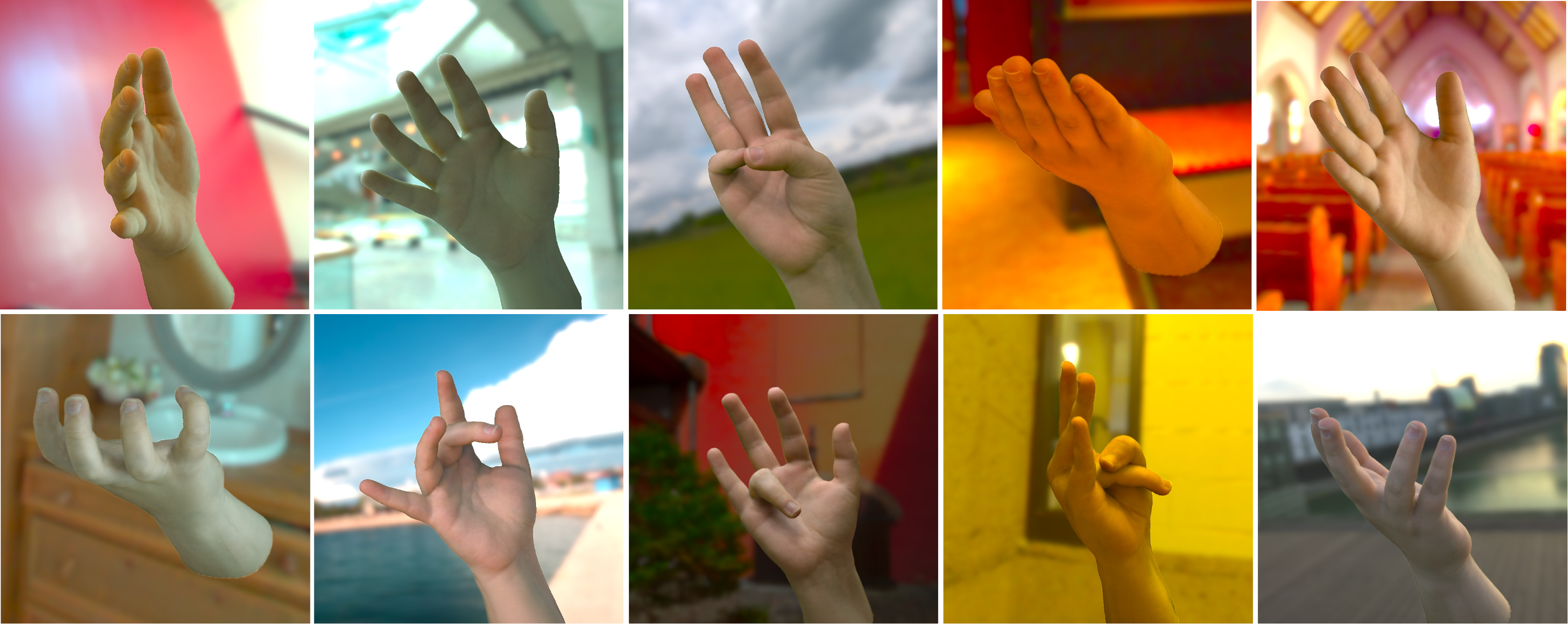}
    \caption{\textbf{Qualitative results of the teacher model with environment map lighting.} We sample the poses from the validation dataset of Subject 1.}
    \label{fig:teacher_envmap_subject1}
\end{figure*}

\begin{figure*}[htb!]
    \centering
    \includegraphics[width=\linewidth]{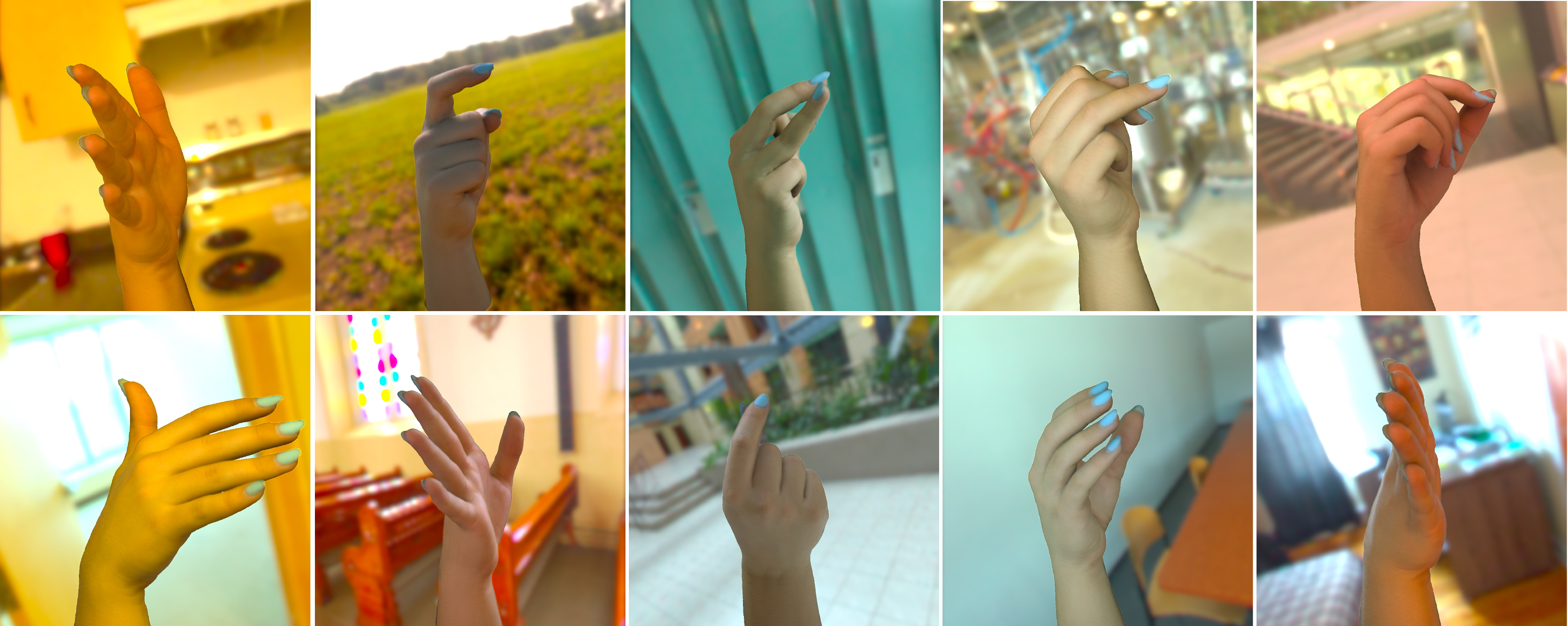}
    \caption{\textbf{Qualitative results of the teacher model with environment map lighting.} We sample the poses from the validation dataset of Subject 2.}
    \label{fig:teacher_envmap_subject2}
\end{figure*}

\begin{figure*}[htb!]
    \centering
    \includegraphics[width=\linewidth]{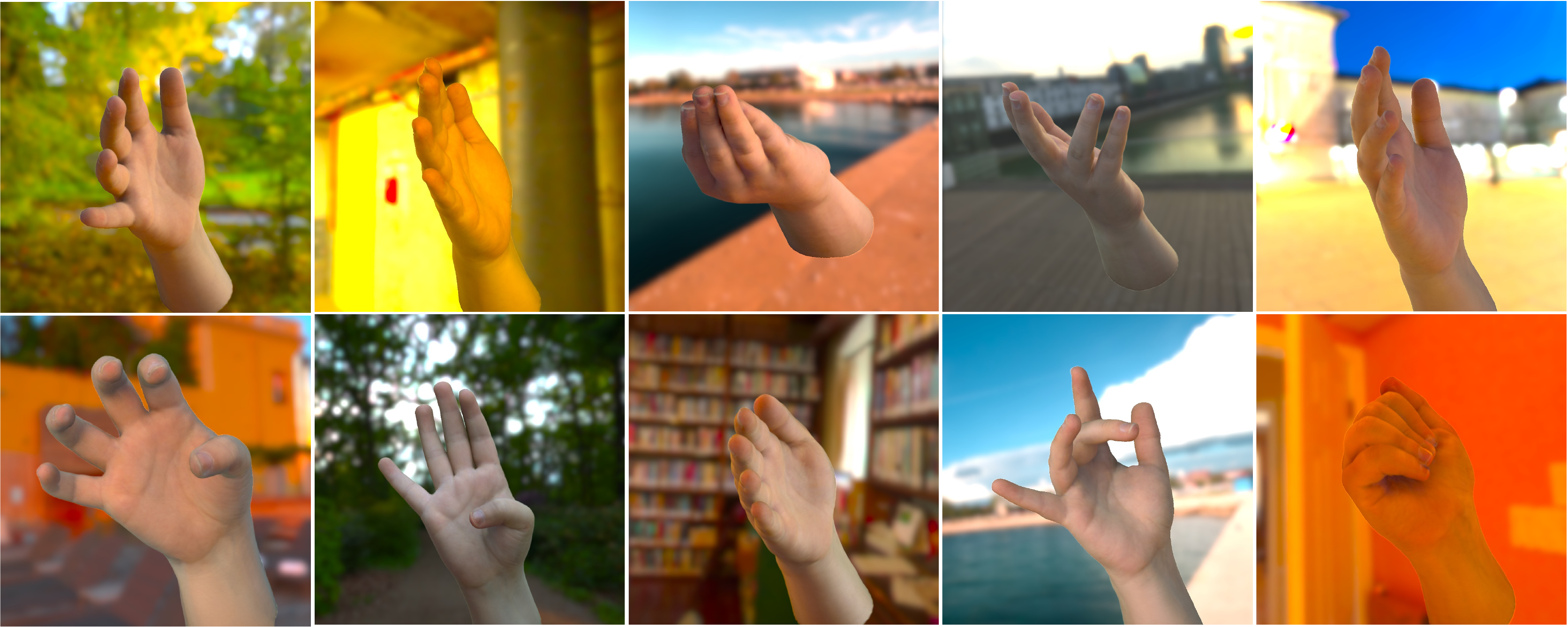}
    \caption{\textbf{Qualitative results of the student model with environment map lighting.} We sample the poses from the validation dataset of Subject 1.}
    \label{fig:student_envmap_subject1}
\end{figure*}

\begin{figure*}[htb!]
    \centering
    \includegraphics[width=\linewidth]{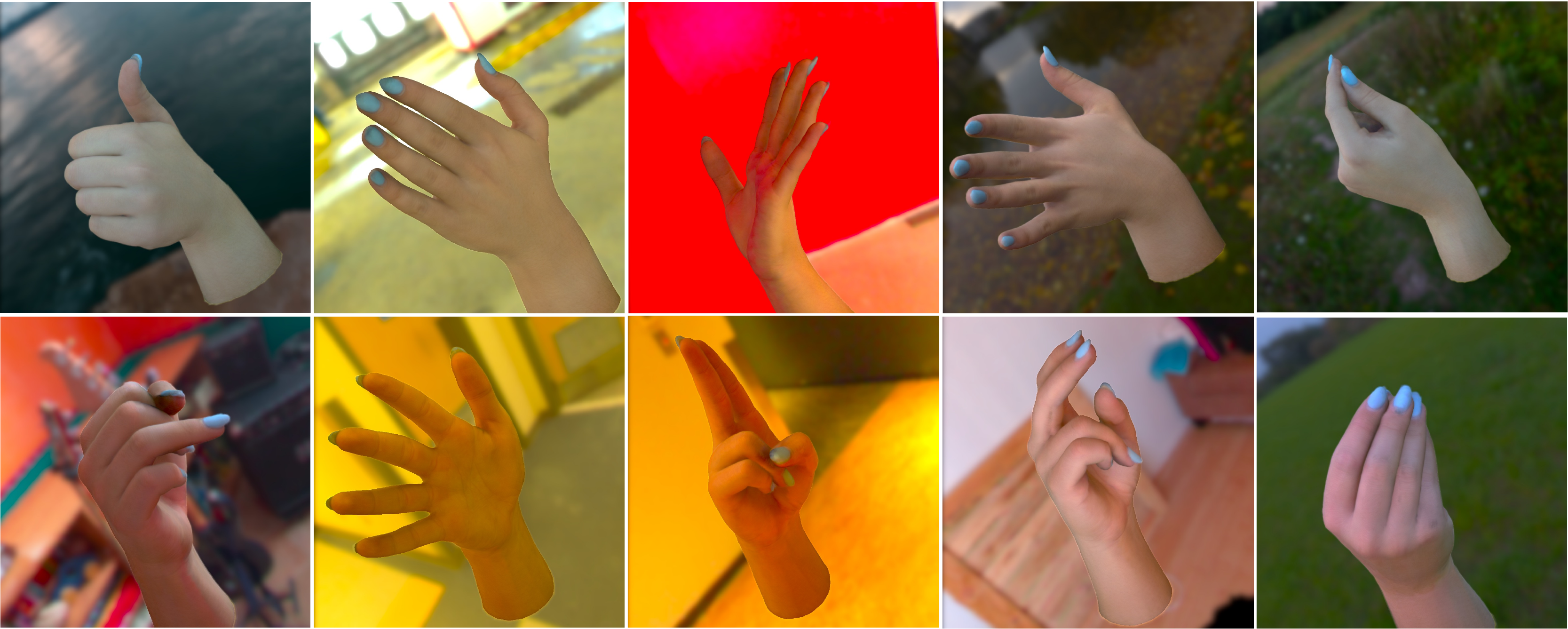}
    \caption{\textbf{Qualitative results of the student model with environment map lighting.} We sample the poses from the validation dataset of Subject 2.}
    \label{fig:student_envmap_subject2}
\end{figure*}

\section{Runtime Analysis}
The student model runtime for a single hand mainly comprises 0.3 ms for joint feature decoding, 2.7 ms for ray tracing, 1.5 ms for grid sampling from 3D to the UV space, and 15.8 ms for texture decoding on NVIDIA V100.

\section{Qualitative Results}
\subsection{Teacher Model}
\Cref{fig:teacher_directional_subject1,fig:teacher_directional_subject2,fig:teacher_nearfield_subject1,fig:teacher_nearfield_subject2,fig:teacher_envmap_subject1,fig:teacher_envmap_subject2} show the results of teacher model with directional lighting, near-field lighting, and environment map rendering for Subject 1 and 2. Note that the hand poses and lighting conditions are randomly sampled from the validation subset. These qualitative results clearly demonstrate the high-fidelity relighting of hand models with various poses and illuminations. Please refer to the supplemental video for animation results.

\subsection{Student Model}
\Cref{fig:student_envmap_subject1,fig:student_envmap_subject2} exhibit the high-fidelity renderings generated by the student model in real-time. Our visibility-aware diffuse and specular features enables the generalization to unseen environment maps and hand poses. Moreover, our method is able to reproduce faithful details such as specularity and soft shadows.

\end{document}